\documentclass{article} 
\usepackage{iclr2027_conference,times}
\usepackage[utf8]{inputenc} 
\usepackage[T1]{fontenc}    
\usepackage{hyperref}       
\usepackage{url}            
\usepackage{booktabs}       
\usepackage{amsfonts}       
\usepackage{nicefrac}       
\usepackage{microtype}      
\usepackage{xcolor}         
\usepackage{amsmath}
\usepackage{cleveref}
\usepackage{multirow}
\usepackage[table]{xcolor}
\usepackage{graphicx}      
\usepackage{wrapfig}

\usepackage{amsmath,amsfonts,bm}

\def\eqref#1{equation~\ref{#1}}

\def\1{\bm{1}}

\DeclareMathAlphabet{\mathsfit}{\encodingdefault}{\sfdefault}{m}{sl}
\SetMathAlphabet{\mathsfit}{bold}{\encodingdefault}{\sfdefault}{bx}{n}

\usepackage{hyperref}
\usepackage{url}

\title{	
Looking Inside LLMs: Small-World Connectivity
as a Signature of Reasoning Performance}

\author{Zheng Huang$^{1}$\thanks{Correspondence to zheng.huang.gr@dartmouth.edu. },
Sansheng Cao$^{1}$,
Enpei Zhang$^{1}$,
Weikang Qiu$^{2}$,
Elynn Chen$^{3}$,\\
\textbf{Xiang Zhang$^{4}$,
Yaoqing Yang$^{1}$,
Rex Ying$^{2}$,
Dawei Zhou$^{5}$,
Yujun Yan$^{1}$}
\\
$^{1}$Dartmouth College \quad
$^{2}$Yale University \quad
$^{3}$New York University \\
$^{4}$UNC Charlotte \quad
$^{5}$Virginia Tech
}\iclrfinalcopy
\begin{document}

\maketitle

\begin{abstract}
Understanding large language model (LLM) reasoning requires looking beyond behavioral performance to examine how reasoning ability is reflected in internal organization. Inspired by neuroscience findings linking higher intelligence to stronger small-world organization in functional brain networks, we investigate small-world connectivity as a structural signature of LLM reasoning. We construct functional graphs from attention-head activation similarities and find that a higher small-world index (SWI), capturing local clustering and short global paths, consistently correlates with better fluid reasoning performance across models and training checkpoints. Since local clustering is central to small-world organization, we further examine how heads important for model performance connect within and across communities. We find that these heads tend to have a larger share of connection weight within their own communities (high core scores) and a more concentrated weight distribution across communities (low bridge scores). These observations motivate the hypothesis that high core and low bridge scores serve as structural indicators of head importance for reasoning capability. We validate this hypothesis through pruning, introducing Small-World Allocation (SWA), a hierarchical sparsity allocation method guided by these scores. Across six LLMs, SWA better preserves small-world organization and model performance than competing allocation strategies, reducing WikiText perplexity by up to 20\%. Together, these findings identify small-world functional connectivity as a measurable signature of LLM reasoning performance, offering a structural perspective that complements behavioral evaluation.
\end{abstract}

\section{Introduction}

Large language models (LLMs) demonstrate increasingly strong performance on reasoning tasks, including mathematical problem solving, logical inference, and reasoning about code \citep{leon2025gpt,liu2025deepseek,wang2024mmlu}. Behavioral benchmarks quantify success on these tasks, but understanding these capabilities also calls for examining how differences in reasoning performance are reflected inside the model. This motivates a central question: Can we identify measurable patterns of internal organization associated with stronger LLM reasoning performance?

Neuroscience offers a precedent for relating intelligence to internal network organization. In functional brain networks, defined by statistical dependencies between regional activities, stronger small-world organization has been associated with higher intelligence \citep{langer2012functional,barbey2018network}. Such organization features high local clustering while maintaining short network paths \citep{latora2001efficient,bassett2017small}. This association motivates us to ask whether analogous functional organization in LLMs provides a structural signature of reasoning performance.

Examining this possibility requires representing relationships expressed through model activity. Existing structural analyses provide complementary but distinct perspectives. Architecture-level graphs relate prescribed connectivity to generalization in vision models \citep{you2020graph}, but primarily reflect model design rather than functional relationships shaped by training. Learned weight spectra capture training-dependent properties associated with generalization and pruning sensitivity \citep{martin2021predicting,hu2025eigenspectrum,lu2024alphapruning}, but characterize weight statistics rather than relationships among component activities. We therefore construct functional graphs with attention heads as nodes and activation similarities as edge weights, capturing relationships expressed in a trained model’s responses to inputs. Using this representation, we first examine whether small-world organization is associated with reasoning performance at the network level. Across models and training checkpoints, we find that a higher small-world index (SWI) consistently correlates with better fluid reasoning performance (\Cref{fig:swi-perf} \& \Cref{fig:swi-perf-cross_model}).

To better understand this network-level association, we examine whether attention heads important for model performance exhibit distinctive connectivity patterns within and across communities, motivated by the local clustering characteristic of small-world organization. We estimate head importance following \citep{michel2019sixteen} and find that more important heads tend to have a larger share of connection weight within their own communities (high core scores) and a more concentrated distribution of weight across communities (low bridge scores; Section~\ref{sec:local_head_analysis}). These observations motivate the hypothesis that high core and low bridge scores serve as structural indicators of head importance for the reasoning ability of LLMs. We validate this hypothesis through pruning: if these scores are informative about head importance, sparsity allocation guided by them should better retain models' reasoning performance. We examine this prediction through Small-World Allocation (SWA), a hierarchical sparsity allocation method guided by core and bridge scores. Across six LLMs, SWA better retains model performance than competing allocation strategies, reducing WikiText perplexity by up to 20\%, while also better preserving small-world organization. These results provide additional evidence that the identified head-level patterns are informative about model performance. Together, our findings characterize small-world functional connectivity as a measurable signature of LLM reasoning and offer a structural
perspective that complements behavioral evaluation. Our contributions are summarized as follows:

\vspace{-0.2cm}
\begin{itemize}
    \item \textbf{New Observation:} To our knowledge, we are the first to show that LLMs' attention head activations exhibit a small-world functional organization. We further show that SWI is correlated with performance on fluid intelligence benchmarks and remains consistent across different LLM architectures.

   \item \textbf{Novel Findings:} Through head importance analysis, we find that heads with low bridge scores and high core scores are significantly more prevalent among heads with greater importance for model performance. This links local connectivity patterns to head importance and motivates the hypothesis that preferentially preserving heads with these structural properties can better retain reasoning performance.
   

    \item \textbf{New Method:} We propose Small-World Allocation (SWA), a hierarchical sparsity allocation method that uses core and bridge scores to guide pruning across layers and attention heads. Experiments show that SWA better preserves small-world organization under pruning and improves language modeling performance by up to $20\%$ over state-of-the-art baselines.
    

\end{itemize}

\section{Preliminaries}
\label{sec:prelim}



\subsection{Small-World Organization of Brain Networks}

Small-world organization describes a network topology characterized by high local clustering and efficient global communication \citep{barbey2018network, rosenberg2020functional}. In neuroscience, functional brain networks are commonly constructed over anatomical regions of interest (ROIs), with nodes representing brain regions and edges representing functional connectivity \citep{langer2012functional}. Stronger small-world structure has been associated with higher human intelligence: higher scores on Raven's Advanced Progressive Matrices \citep{bors2003effect, raven1962advanced} are linked to higher clustering coefficients and shorter communication path lengths \citep{langer2012functional}.

More generally, given a graph $\mathcal{G}=(\mathcal{V},\mathcal{E})$, where $\mathcal{V}$ is the node set and $\mathcal{E}$ is the edge set, let $C(\mathcal{G})$ denote its average clustering coefficient \citep{watts1998collective}, defined as
\begin{equation}
C(\mathcal{G}) =
\frac{1}{|\mathcal{V}|}
\sum_{i\in\mathcal{V}} C_i,
\qquad
C_i =
\frac{2e_i}{d_i(d_i-1)},
\end{equation}
where $d_i$ is the degree of node $i$, $e_i$ is the number of edges among the neighbors of node $i$, and we set $C_i=0$ when $d_i<2$. We let $E(\mathcal{G})$ denote its global efficiency, which captures efficient communication through short paths \citep{latora2001efficient}:
\begin{equation}
E(\mathcal{G}) =
\frac{1}{|\mathcal{V}|(|\mathcal{V}|-1)}
\sum_{\substack{i,j\in\mathcal{V}\\ i\neq j}}
\frac{1}{\operatorname{dist}(i,j)},
\end{equation}
where $\operatorname{dist}(i,j)$ is the shortest-path distance between nodes $i$ and $j$, and disconnected node pairs contribute zero efficiency. To assess small-world organization, $\mathcal{G}$ is compared against degree-preserving randomized graphs generated by edge swaps \citep{hosseini2013influence}. Let $C_{\mathrm{rand}}$ and $E_{\mathrm{rand}}$ denote the average clustering coefficient and global efficiency of these randomized graphs. The small-world index of $\mathcal{G}$ is then defined as
\begin{equation}
\mathrm{SWI}(\mathcal{G}) =
\frac{C(\mathcal{G})}{C_{\mathrm{rand}}}
\cdot
\frac{E(\mathcal{G})}{E_{\mathrm{rand}}}.
\label{eq:swi}
\end{equation}
A larger SWI indicates stronger local clustering and more efficient global communication relative to degree-matched random graphs.

\subsection{Representation Similarity}
\label{sec:svcca_prelim}


Singular Vector Canonical Correlation Analysis (SVCCA)~\citet{raghu2017svcca} is used to measure similarity between vector-valued representations. Given two mean-centered representation matrices
$\mathbf{X}\in\mathbb{R}^{m\times D_x}$ and
$\mathbf{Y}\in\mathbb{R}^{m\times D_y}$ over the same $m$ examples, SVCCA first applies singular value decomposition:
\begin{equation}
\mathbf{X} = \mathbf{U}_X \mathbf{S}_X \mathbf{V}_X^\top,
\qquad
\mathbf{Y} = \mathbf{U}_Y \mathbf{S}_Y \mathbf{V}_Y^\top .
\end{equation}

It then retains the top singular directions explaining a high proportion of variance, e.g., $99\%$. Let $p_x$ and $p_y$ be the numbers of retained directions for $\mathbf{X}$ and $\mathbf{Y}$, respectively. The representations are then projected as

\begin{equation}
\mathbf{X}' = \mathbf{X}\mathbf{V}_X^{(p_x)}, 
\qquad
\mathbf{Y}' = \mathbf{Y}\mathbf{V}_Y^{(p_y)} ,
\end{equation}

where $\mathbf{X}'\in\mathbb{R}^{m\times p_x}$ and
$\mathbf{Y}'\in\mathbb{R}^{m\times p_y}$. Here, $\mathbf{V}_X^{(p_x)}$ and $\mathbf{V}_Y^{(p_y)}$
denote the matrices formed by the first $p_x$ and $p_y$ columns of
$\mathbf{V}_X$ and $\mathbf{V}_Y$, respectively.

Canonical Correlation Analysis (CCA)~\citep{thompson2000canonical} then finds projection vectors $\mathbf{w}_x$ and $\mathbf{w}_y$ that maximize the correlation between the projected representations:
$
\rho =
\max_{\mathbf{w}_x, \mathbf{w}_y}
\frac{
\mathbf{w}_x^\top \boldsymbol{\Sigma}_{X'Y'} \mathbf{w}_y
}{
\sqrt{
(\mathbf{w}_x^\top \boldsymbol{\Sigma}_{X'X'} \mathbf{w}_x)
(\mathbf{w}_y^\top \boldsymbol{\Sigma}_{Y'Y'} \mathbf{w}_y)
}
}.
$
Here, $\boldsymbol{\Sigma}_{X'X'}$ and $\boldsymbol{\Sigma}_{Y'Y'}$ denote the covariance matrices of
$\mathbf{X}'$ and $\mathbf{Y}'$, and $\boldsymbol{\Sigma}_{X'Y'}$ denotes their cross-covariance matrix. This produces a set of canonical correlations
$\{\rho_q\}_{q=1}^{r}$, where $r=\min(p_x,p_y)$.
The SVCCA similarity is defined as their mean:
\begin{equation}
\mathrm{SVCCA}(\mathbf{X},\mathbf{Y})
=
\frac{1}{r}\sum_{q=1}^{r}\rho_q .
\end{equation}
Higher SVCCA values indicate stronger alignment between representational subspaces. In this work, we use a variant of SVCCA that replaces mean aggregation with the sum of Fisher
$z$-transformed canonical correlations, as detailed
in~\Cref{eq:fisher_similarity}.
\section{Key Findings}
\label{key_findings}
\begin{figure}
    \centering
    \includegraphics[width=0.85\linewidth]{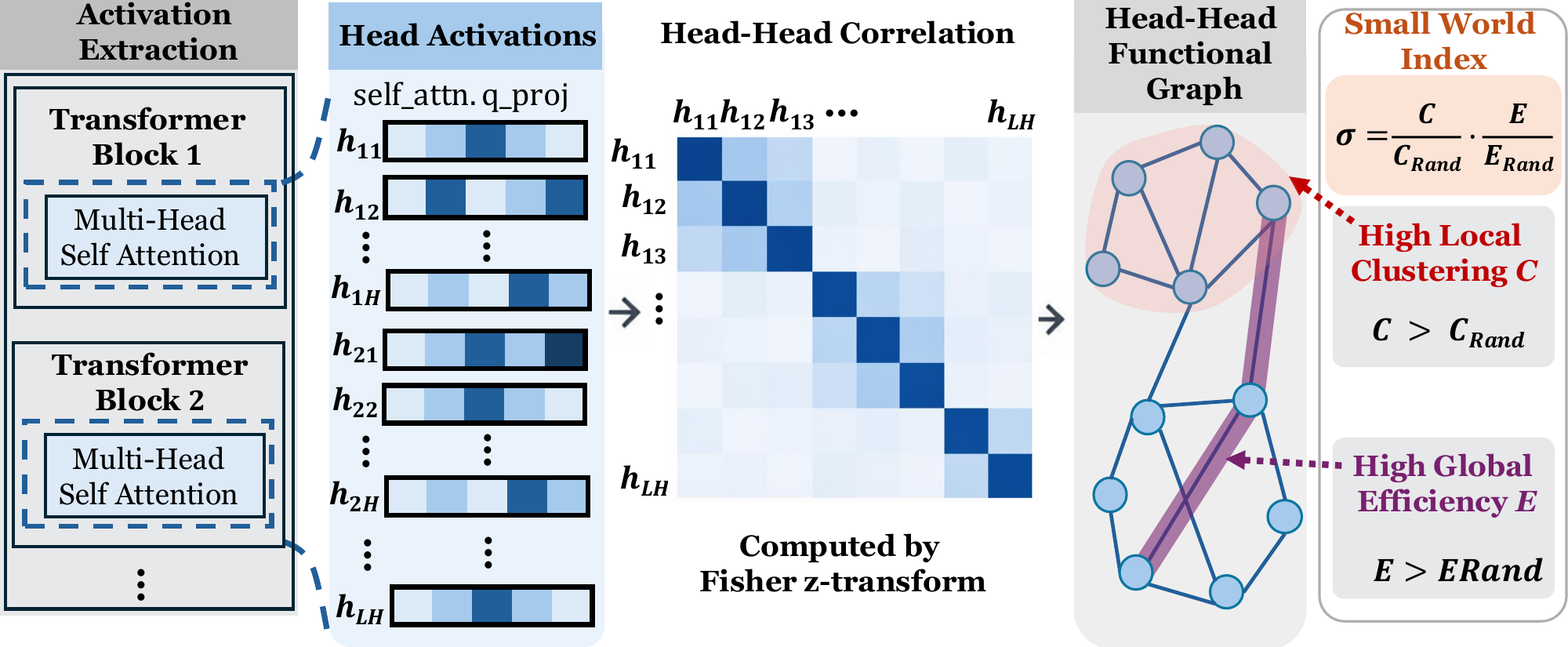}
    \caption{\small We extract activations from attention heads across Transformer layers and compute pairwise similarity to construct a head-level functional graph. The resulting graph exhibits small-world structure, characterized by high clustering and efficient short-path communication. We show that this structural property correlates with model capability on fluid intelligence benchmarks.}
    \label{fig:process}
\end{figure}

In this section, we present our framework for constructing functional graphs of attention heads and quantifying their small-world organization. We first show how we construct an attention-head functional graph and compute its SWI (\Cref{sec:constructing-head}). Second, we show that SWI is associated with LLM performance on fluid intelligence benchmarks (\Cref{sec:quantify_swi}). The process is shown in~\Cref{fig:process}.

\subsection{Constructing Functional Graphs of Attention Heads}
\label{sec:constructing-head}

We construct attention-head functional graphs by adapting the standard procedure used to build functional brain networks. In brain networks, anatomical regions of interest (ROIs) serve as nodes, while edges capture statistical dependencies between their activity time series; thresholding the resulting connectivity matrix produces the functional graph. Following the same construction, we treat attention heads as nodes and define their functional connectivity using activation-based similarity. We then threshold the resulting head-similarity matrix to obtain the attention-head functional graph.

\paragraph{Activation Collection.}
We consider a transformer-based LLM with $L$ layers and
$H$ attention heads per layer, yielding $M=LH$ heads
in total. We denote the set of all attention heads by
$
\mathcal{H} = \{h_{\ell,a} \mid \ell = 1,\ldots,L,\;
a = 1,\ldots,H\},
$
where $h_{\ell,a}$ denotes the $a$-th attention head
in layer $\ell$. We sample $N$ questions
$\{x^{(n)}\}_{n=1}^{N}$ from a given dataset
(e.g., DRE-Bench~\citep{yang2025truly}) for graph
construction. These questions are excluded from
downstream evaluations. For each question $x^{(n)}$, let $T_n$ denote the number of tokens generated in its answer. We record the query activations of all self-attention heads at each generated answer token.
This yields an activation tensor:
$
\mathbf{A}^{(n)} \in \mathbb{R}^{M \times T_n \times d},
$
where $d$ is the head dimension and $\mathbf{A}^{(n)}_{m,t,:}$ denotes the query activation of head $m$ at the $t$-th generated answer token. For each head and question, we average its token-level activations over the generated answer tokens:
$
\mathbf{z}^{(n)}_m =
\frac{1}{T_n}\sum_{t=1}^{T_n}\mathbf{A}^{(n)}_{m,t,:}, \;\mathbf{z}^{(n)}_m \in \mathbb{R}^{d}.
$
Stacking these vectors over all $N$ questions gives the activation matrix for head $m$:
$
\mathbf{X}_m =
\left[
\mathbf{z}^{(1)}_m;
\mathbf{z}^{(2)}_m;
\dots;
\mathbf{z}^{(N)}_m
\right]
\in \mathbb{R}^{N\times d}.
$
Each row of $\mathbf{X}_m$ corresponds to the query activation of head $m$ for one question.

\paragraph{Head Similarity and Graph Construction.}

In functional brain networks, connectivity is often estimated using Pearson correlation between scalar ROI activity time series~\citep{biswal1995functional,fox2005human}. Attention heads instead produce vector-valued activations, so we use a variant of SVCCA that Fisher (z)-transforms and aggregates canonical correlations to quantify functional similarity between heads. Further justification is provided in Appendix~\ref{app:svcca_justification}.

We treat the activation matrix $\mathbf{X}_m$ of each head
as its representation across the sampled questions.
For each pair of distinct heads $(i,j)$, we apply the
SVD and CCA procedure described in~\Cref{sec:svcca_prelim}
to obtain canonical correlations $\{\rho_q\}_{q=1}^{r}$.
We define their similarity by summing the transformed
coefficients:
\begin{equation}
S_{ij}
=
S(\mathbf{X}_i,\mathbf{X}_j)
=
\sum_{q=1}^{r}\operatorname{atanh}(\tilde{\rho}_q)
=
\frac{1}{2}\sum_{q=1}^{r}
\log\left(
\frac{1+\tilde{\rho}_q}{1-\tilde{\rho}_q}
\right).
\label{eq:fisher_similarity}
\end{equation}
Here, $\tilde{\rho}_q=\min(\rho_q,1-\epsilon_{\mathrm{clip}})$,
where $\epsilon_{\mathrm{clip}}\in(0,1)$ is a small constant
that prevents divergence at $\rho_q=1$.
The Fisher $z$-transform emphasizes strongly correlated
canonical directions. This yields a symmetric head-similarity
matrix $\mathbf{S}\in\mathbb{R}^{M\times M}$,
with $\mathbf{S}_{ii}=1$.
A larger $\mathbf{S}_{ij}$ indicates greater aggregate covariation
between the two heads' query representations across the
same questions, reflecting more closely aligned activation
patterns in their retained subspaces.

We then construct an undirected functional graph by
retaining the strongest pairwise similarities at a fixed
edge density. Let $\mathcal{P}$ denote the set of all
unordered pairs of distinct heads. For a target edge
density $\delta$, we rank these pairs by $S_{ij}$ and
retain the top $K_\delta=\lfloor\delta|\mathcal{P}|\rfloor$
pairs as edges. The resulting graph is
$\mathcal{G}_\delta=(\mathcal{V},\mathcal{E}_\delta)$,
where $\mathcal{V}=\{1,\ldots,M\}$ indexes the attention
heads and $\mathcal{E}_\delta$ contains the retained pairs.

\begin{figure}[t]
    \centering
    \begin{minipage}[t]{0.48\textwidth}
        \centering
        \vspace{0pt}
        \includegraphics[width=\linewidth]
        {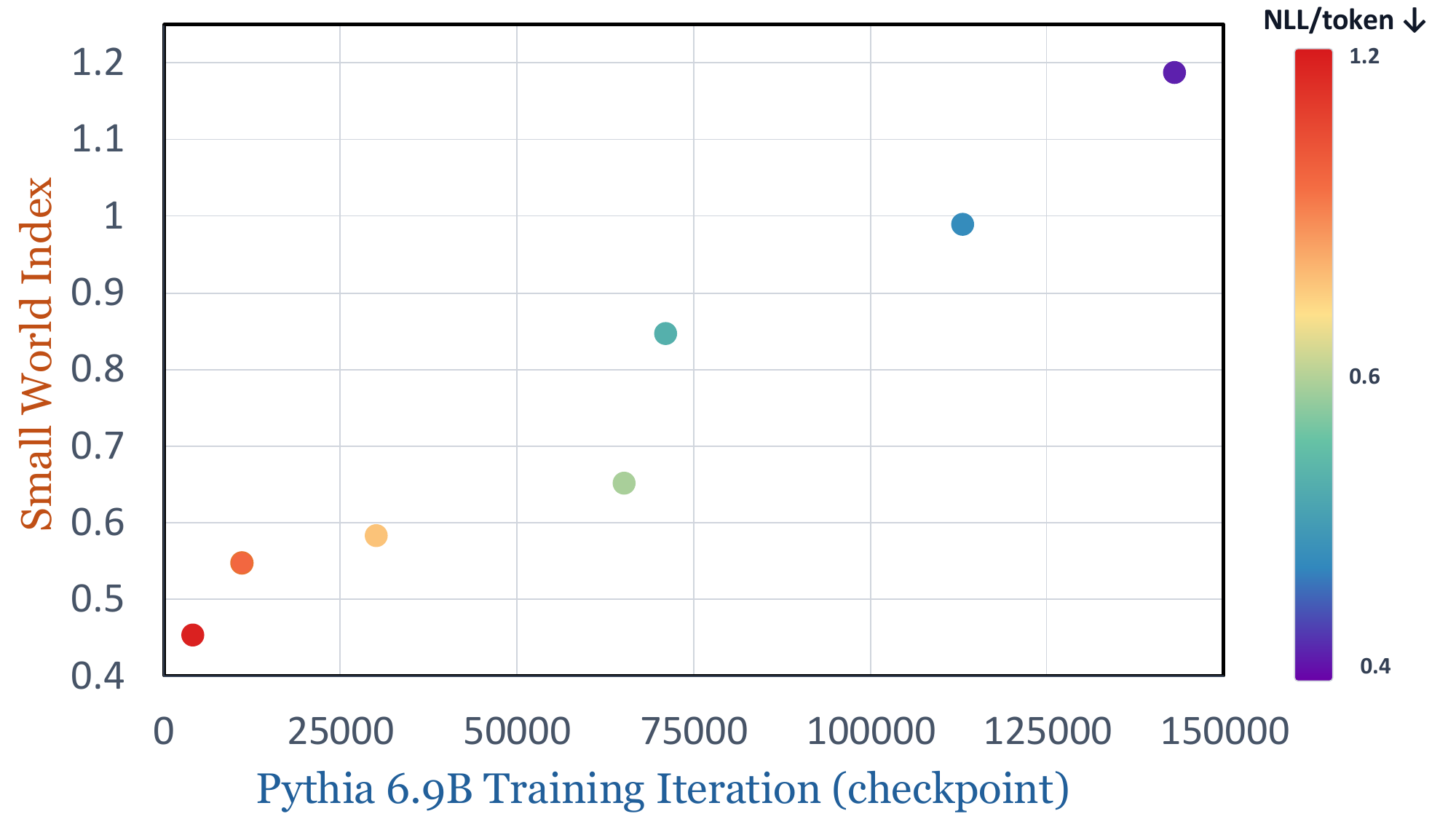}
        \caption{\small
        Across Pythia-6.9B training checkpoints (x-axis),
        a higher small-world index (SWI; y-axis) is
        consistently associated with better fluid reasoning
        performance on DRE-Bench \citep{yang2025truly},
        as reflected by lower negative log-likelihood
        (NLL) per answer token (color).
        }
        \label{fig:swi-perf}
    \end{minipage}
    \hfill
    \begin{minipage}[t]{0.48\textwidth}
        \centering
        \vspace{0pt}
        \includegraphics[width=\linewidth]
        {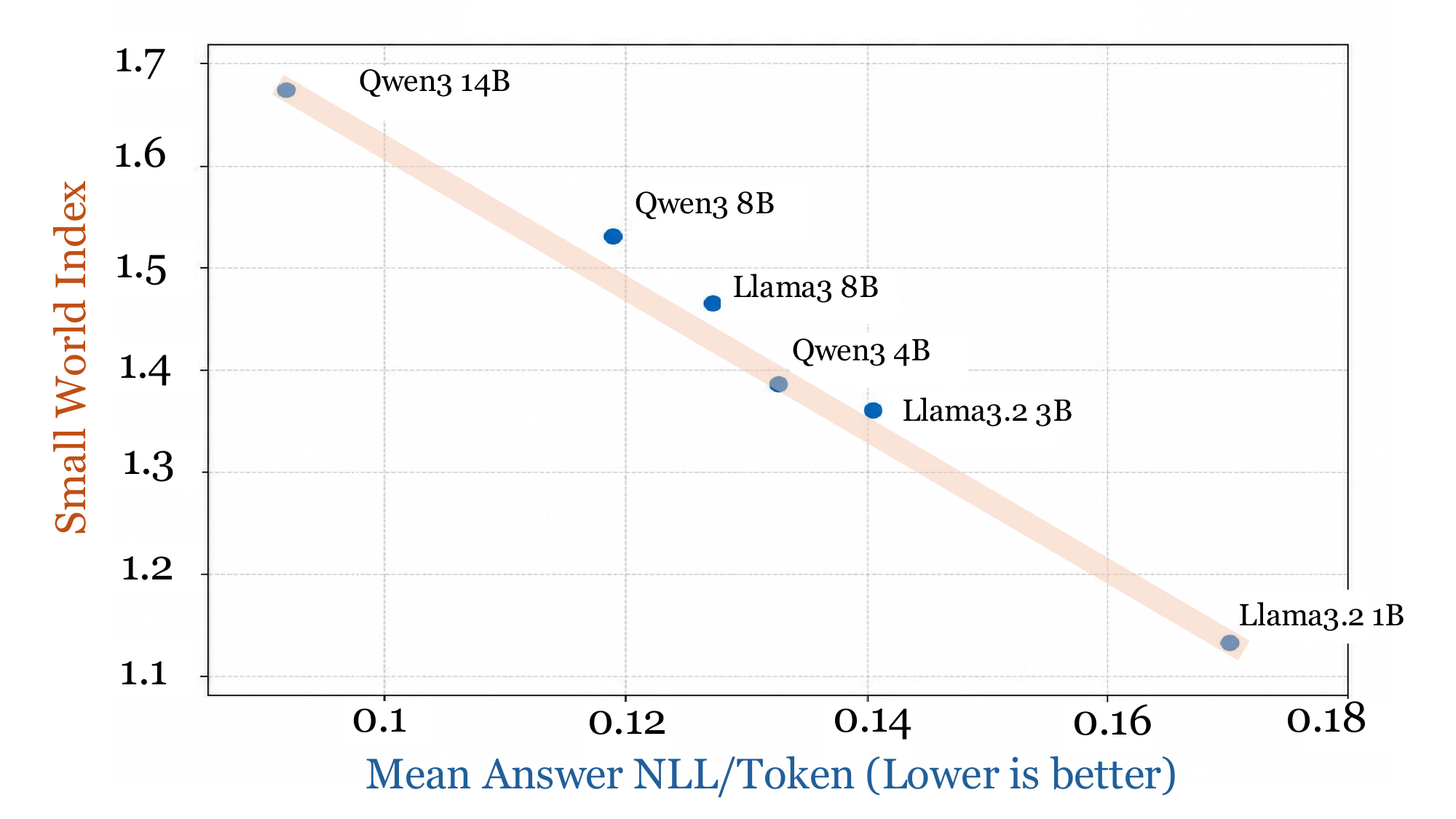}
        \caption{\small
        Across different LLMs, a higher small-world index
        (SWI; y-axis) is consistently associated with
        better fluid reasoning performance, measured
        by lower negative log-likelihood (NLL) per
        answer token on DRE-Bench
        \citep{yang2025truly} (x-axis).
        }
        \label{fig:swi-perf-cross_model}
    \end{minipage}
\end{figure}

\vspace{-0.3cm}

\subsection{Connection of Small-world Index to Fluid Intelligence}
\label{sec:quantify_swi}
We find that \textbf{stronger small-world organization in
head--head functional graphs is associated with better
fluid reasoning performance}, both across training
checkpoints of the same model and across different models.
For each model or checkpoint, we compute the SWI of
$\mathcal{G}_\delta$ constructed on DRE-Bench
using~\Cref{eq:swi}. We measure performance using the
mean per-token negative log-likelihood over the answer
tokens (NLL/token), where lower values indicate better
performance.

As shown in Figure~\ref{fig:swi-perf}, later checkpoints
of Pythia-6.9B generally exhibit higher SWI and lower NLL
than earlier checkpoints. Similarly,
Figure~\ref{fig:swi-perf-cross_model} shows that models
with lower NLL consistently exhibit higher SWI. Together, these observations suggest that
the SWI--performance association holds both within a
model over training and across models. The same trend can be observed in graphs constructed using MMLU \citep{hendrycks2020measuring}.
Experimental details are provided
in~\Cref{sec:exp_setup}, with detailed numerical results
in Appendix~\ref{app:add_swi_fluid}.

\vspace{-0.2cm}




\section{Exploring the SWI--Performance Association through Local Structure}
\label{sec:core-bridge}

Building on the association between SWI and fluid reasoning
performance observed in~\Cref{sec:quantify_swi}, we explore
local connectivity patterns that help explain this
association.
We first identify attention heads that are important for
model performance and examine their local connectivity
patterns (Section~\ref{sec:local_head_analysis}).
Our findings motivate the hypothesis that strong
within-community connectivity and a concentrated distribution
of connection weight across communities play an important
role in supporting LLM reasoning performance.
To test this hypothesis, we use pruning to examine whether retaining heads with these properties
better preserves reasoning performance (Section~\ref{sec:pruning}).

\begin{table}[t]
\centering
\caption{
Prevalence of low bridge scores ($b$) and high core scores
($c$) in Llama-3.2-3B on DRE-Bench, comparing the top 25\% of heads ranked by gradient-based
importance with the remaining 75\%. Low bridge and high core scores are defined as
$b\leq\operatorname{median}(b)$ and
$c>\operatorname{median}(c)$, respectively, with medians
computed across all heads. Two-sided $p$-values test the null hypothesis of equal prevalence between the two groups and are adjusted for multiple comparisons using Holm's method~\citep{holm1979simple}. Smaller $p$-values indicate stronger evidence that the two groups differ in the proportion of heads exhibiting the corresponding structural feature.
}

\label{tab:importance_groups}
\small
\setlength{\tabcolsep}{10pt}
\renewcommand{\arraystretch}{1.15}
\begin{tabular}{@{}lccc@{}}
\toprule
Head property & Top 25\% & Remaining 75\% & $p$-value \\
\midrule
Low Bridge Score 
& \textbf{89.3\%} & 36.9\% & $3.61 \times 10^{-34}$ \\
High Core Score 
& \textbf{49.4\%} & 23.8\% & $5.95 \times 10^{-9}$ \\
\bottomrule
\end{tabular}
\end{table}


\subsection{Local Connectivity Patterns of Important Heads}
\label{sec:local_head_analysis}

In this section, we examine whether heads that are
important for model performance exhibit local
connectivity patterns.
We estimate head importance, characterize connectivity
using core and bridge scores, and compare these structural
properties between important heads and the remaining heads.

Following~\citet{michel2019sixteen}, 
we scale each head's
output by a scalar multiplier $g_i$ and evaluate
gradients with all multipliers set to one. Head importance is the absolute derivative of the mean
answer-token negative log-likelihood with respect to $g_i$,
averaged across test examples.


To characterize local connectivity, we apply Louvain
community detection~\citep{blondel2008fast} to
$\mathcal{G}_\delta$ to obtain communities
$\mathcal{C}_\delta=\{C_1,\ldots,C_K\}$.
Let $C(i)$ denote the community containing head $i$,
and set $S_{ij}=0$ for non-edge pairs in
$\mathcal{G}_\delta$.
We define the \emph{core score} as
$
\mathrm{Core}_i =
\frac{\sum_{j\in C(i)} S_{ij}}
{\sum_{j\in\mathcal{V}} S_{ij}+\varepsilon},
$
where $\varepsilon$ is a small positive constant used
for numerical stability.
A higher core score indicates that a larger proportion
of a head's total connection weight lies within its own
community.
We define the \emph{bridge score} using the weighted
participation coefficient~\citep{pedersen2017spontaneous}:
$
\mathrm{Bridge}_i =
1-\sum_{k=1}^{K}
\left(\frac{s_{i,k}}{s_i+\varepsilon}\right)^2,
$
where
$s_{i,k}=\sum_{j\in C_k}S_{ij}$
and
$s_i=\sum_{j\in\mathcal{V}}S_{ij}$.
A higher bridge score indicates that connection weight
is distributed more evenly across communities, whereas
a lower score indicates greater concentration in fewer
communities.

As shown in Table~\ref{tab:importance_groups}, low bridge
scores and high core scores are each significantly more
prevalent among important heads than among the remaining
heads. This motivates our hypothesis that attention heads with
high core scores and low bridge scores play an important
role in LLM reasoning.

\subsection{Validating Hypothesis Via Pruning}
\label{sec:pruning}
To test whether the identified attention heads with these structural properties
support LLM reasoning performance, we use pruning to
examine whether retaining heads with
high core scores and low bridge scores better preserves
performance.
Specifically, we propose Small-World Allocation (SWA),
a hierarchical sparsity allocation method that uses
core and bridge scores to allocate sparsity across
layers and attention heads, assigning lower sparsity
to heads with these properties.

SWA defines pruning priority scores at both the layer and head levels. Here, each layer corresponds to a transformer block containing self-attention and feed-forward modules. Layer-level scores allocate sparsity across these blocks, while head-level scores determine the pruning ratios of individual attention
heads within each block. This hierarchical design accounts for structural differences both across layers
and among heads within a layer.

At the head level, we combine core and bridge scores into a pruning priority score. For each head $i$, we first convert its bridge and core scores into percentile ranks over all heads, denoted by $\tilde{B}_i \in [0,1]$ and $\tilde{C}_i \in [0,1]$, where larger values indicate stronger structural roles. We define the priority score as
\begin{equation}
p_i = \tilde{B}_i(1-\tilde{C}_i).
\end{equation}

Heads with low bridge scores or high core scores receive smaller $p_i$ values and are assigned lower sparsity, preserving more of their weights. At the layer level, we aggregate head-level pruning priority scores within each layer. Let $\mathcal{H}_\ell$ denote the set of heads in layer $\ell$, we have
$
p_\ell =
\frac{1}{|\mathcal{H}_\ell|}
\sum_{i\in \mathcal{H}_\ell}
p_i .
$

A smaller score $p_\ell$ indicates that layer $\ell$ contains more structurally important heads and should receive a lower sparsity budget, while a larger $p_\ell$ indicates higher layer-level pruning priority.

SWA assigns pruning budgets in proportion to the priority scores through a hierarchical allocation procedure. Layer-level scores first determine the sparsity budget for each transformer block, including all modules within the block; head-level scores then distribute the attention sparsity across individual heads. Following~\citet{hu2025eigenspectrum}, we normalize the scores, convert them into relative pruning ratios, and rescale the ratios so that the overall sparsity matches the target global sparsity.
\section{Experiments}
\begin{table}[t]
\centering
\small
\setlength{\tabcolsep}{3.0pt}
\caption{\small 
We report the small-world index (SWI $\uparrow$) of head-head functional graphs after applying SparseGPT and its variants at sparsity 0.5. 
 SWA (Head) and SWA (Layer) denote head-only and layer-only allocation strategies, respectively. 
\textbf{Higher SWI indicates better preservation} of small-world functional organization under pruning. Results are averaged over three seeds and reported as mean$_{\pm \text{std}}$.
The best result for each model is \colorbox{gray!15}{\textbf{bolded}}.
}
\resizebox{0.9\linewidth}{!}{

\begin{tabular}{lcccccc}
\toprule
\textbf{Model} 
& \textbf{SparseGPT} 
& \textbf{+ FARMS} 
& \textbf{+ ATP} 
& \textbf{+ SWA (Head)} 
& \textbf{+ SWA (Layer)} 
& \textbf{+ SWA} \\
\midrule
Qwen3 14B     & 0.8000$_{\scriptstyle \pm 0.76}$ & 0.8031$_{\scriptstyle \pm 0.09}$ & 0.8797$_{\scriptstyle \pm 0.37}$ & 0.9867$_{\scriptstyle \pm 0.23}$ & 1.057$_{\scriptstyle \pm 0.14}$ & \colorbox{gray!15}{\textbf{1.1386}$_{\scriptstyle \pm 0.40}$} \\
Qwen3 8B      & 0.6509$_{\scriptstyle \pm 0.64}$ & 0.7982$_{\scriptstyle \pm 0.90}$ & 0.8432$_{\scriptstyle \pm 0.30}$ & 0.8639$_{\scriptstyle \pm 0.22}$ & 0.9053$_{\scriptstyle \pm 0.15}$ & \colorbox{gray!15}{\textbf{0.9196}$_{\scriptstyle \pm 0.44}$} \\
Qwen3 4B      & 0.5400$_{\scriptstyle \pm 0.10}$ & 0.5516$_{\scriptstyle \pm 0.81}$ & 0.6555$_{\scriptstyle \pm 0.14}$ & 0.6449$_{\scriptstyle \pm 0.06}$ & 0.7107$_{\scriptstyle \pm 0.25}$ & \colorbox{gray!15}{\textbf{0.7212}$_{\scriptstyle \pm 0.58}$} \\
\midrule
Llama3 8B    & 0.8085$_{\scriptstyle \pm 0.05}$ & 0.8229$_{\scriptstyle \pm 0.28}$ & 0.8300$_{\scriptstyle \pm 0.68}$ & 0.8634$_{\scriptstyle \pm 0.64}$ & 0.8943$_{\scriptstyle \pm 0.22}$ & \colorbox{gray!15}{\textbf{0.9040}$_{\scriptstyle \pm 0.45}$} \\
Llama3.2 3B  & 0.8372$_{\scriptstyle \pm 0.21}$ & 0.8380$_{\scriptstyle \pm 0.27}$ & 0.8530$_{\scriptstyle \pm 0.14}$ & 0.8468$_{\scriptstyle \pm 0.26}$ & 0.8777$_{\scriptstyle \pm 0.09}$ & \colorbox{gray!15}{\textbf{0.8867}$_{\scriptstyle \pm 0.24}$} \\
Llama3.2 1B  & 0.5138$_{\scriptstyle \pm 0.50}$ & 0.5408$_{\scriptstyle \pm 0.40}$ & 0.5567$_{\scriptstyle \pm 0.54}$ & 0.5943$_{\scriptstyle \pm 0.93}$ & 0.6641$_{\scriptstyle \pm 0.67}$ & \colorbox{gray!15}{\textbf{0.6740}$_{\scriptstyle \pm 0.43}$} \\
\bottomrule
\end{tabular}
}
\label{tab:swi_pruning_preservation}

\end{table}
\begin{table*}[t]
\centering
\scriptsize
\setlength{\tabcolsep}{2.0pt}
\caption{\small  Comparison of \textbf{WikiText Perplexity ($\downarrow$)}. We evaluate Base pruning methods (Wanda and SparseGPT) and their variants augmented with FARMS, ATP, and SWA at sparsity 0.5/0.7. Results are averaged over three seeds and reported as mean$_{\pm \text{std}}$. The best result in each setting is \colorbox{gray!15}{\textbf{bolded}}. SWA reduces perplexity and achieves the best values across all the models and sparsity levels.}
\begin{tabular}{ccc|llll|llll}
\toprule
\multirow{2}{*}{\textbf{Model}} & \multirow{2}{*}{\textbf{Size}} & \multirow{2}{*}{\textbf{Sparsity}} & \multicolumn{4}{c|}{\textbf{Wanda}} & \multicolumn{4}{c}{\textbf{SparseGPT}} \\
 &  &  & Base & + FARMS & + ATP & \textbf{+ SWA} & Base & + FARMS & + ATP & \textbf{+ SWA} \\
\midrule
\multirow{4}{*}{\shortstack{Llama 3.2}} & \multirow{2}{*}{1B} & 0.5 & 36.25$_{\scriptstyle \pm 0.18}$ & 33.90$_{\scriptstyle \pm 0.02}$ & 24.95$_{\scriptstyle \pm 0.04}$ & \cellcolor{gray!15}\textbf{21.92$_{\scriptstyle \pm 0.05}$} & 21.54$_{\scriptstyle \pm 0.29}$ & 21.14$_{\scriptstyle \pm 0.56}$ & 20.56$_{\scriptstyle \pm 0.34}$ & \cellcolor{gray!15}\textbf{20.47$_{\scriptstyle \pm 0.15}$} \\
 &  & 0.7 & 415.60$_{\scriptstyle \pm 32.1}$& 380.20$_{\scriptstyle \pm 20.1}$& 264.10$_{\scriptstyle \pm 21.7}$& \cellcolor{gray!15}\textbf{256.67$_{\scriptstyle \pm 18.00}$} & 157.20$_{\scriptstyle \pm 3.2}$& 136.80$_{\scriptstyle \pm 5.3}$& 106.60$_{\scriptstyle \pm 3.2}$& \cellcolor{gray!15}\textbf{105.00$_{\scriptstyle \pm 3.6}$} \\
 & \multirow{2}{*}{3B} & 0.5 & 19.71$_{\scriptstyle \pm 0.02}$ & 16.45$_{\scriptstyle \pm 0.09}$ & 14.42$_{\scriptstyle \pm 0.02}$ & \cellcolor{gray!15}\textbf{13.32$_{\scriptstyle \pm 0.12}$} & 23.98$_{\scriptstyle \pm 2.15}$ & 15.92$_{\scriptstyle \pm 1.85}$ & 13.53$_{\scriptstyle \pm 0.08}$ & \cellcolor{gray!15}\textbf{11.27$_{\scriptstyle \pm 0.98}$} \\
 &  & 0.7 & 177.80$_{\scriptstyle \pm 5.6}$ & 132.6$_{\scriptstyle \pm 5.3}$ & 125.1$_{\scriptstyle \pm 0.8}$ & \cellcolor{gray!15}\textbf{118.67$_{\scriptstyle \pm 2.32}$} & 70.70$_{\scriptstyle \pm 5.07}$ & 67.67$_{\scriptstyle \pm 2.93}$ & 51.51$_{\scriptstyle \pm 3.83}$ & \cellcolor{gray!15}\textbf{49.30$_{\scriptstyle \pm 0.93}$} \\
\multirow{2}{*}{\shortstack{Llama 3}} & \multirow{2}{*}{8B} & 0.5 & 11.86$_{\scriptstyle \pm 0.02}$ & 11.12$_{\scriptstyle \pm 0.03}$ & 11.50$_{\scriptstyle \pm 0.05}$ & \cellcolor{gray!15}\textbf{10.14$_{\scriptstyle \pm 0.03}$} & 11.69$_{\scriptstyle \pm 0.13}$ & 11.02$_{\scriptstyle \pm 0.87}$ & 10.72$_{\scriptstyle \pm 0.15}$ & \cellcolor{gray!15}\textbf{10.15$_{\scriptstyle \pm 0.08}$} \\
 &  & 0.7 & 140.50$_{\scriptstyle \pm 2.6}$ & 90.02$_{\scriptstyle \pm 3.00}$ & 85.93$_{\scriptstyle \pm 0.82}$ & \cellcolor{gray!15}\textbf{82.36$_{\scriptstyle \pm 0.20}$} & 55.94$_{\scriptstyle \pm 3.23}$ & 52.55$_{\scriptstyle \pm 1.98}$ & 40.40$_{\scriptstyle \pm 1.94}$ & \cellcolor{gray!15}\textbf{38.93$_{\scriptstyle \pm 1.29}$} \\
\midrule
\multirow{6}{*}{\shortstack{Qwen 3}} & \multirow{2}{*}{4B} & 0.5 & 21.91$_{\scriptstyle \pm 2.08}$ & 19.69$_{\scriptstyle \pm 0.16}$ & 18.88$_{\scriptstyle \pm 0.03}$ & \cellcolor{gray!15}\textbf{17.30$_{\scriptstyle \pm 0.60}$} & 20.52$_{\scriptstyle \pm 0.43}$ & 19.58$_{\scriptstyle \pm 0.41}$ & 20.44$_{\scriptstyle \pm 0.45}$ & \cellcolor{gray!15}\textbf{19.30$_{\scriptstyle \pm 0.77}$} \\
 &  & 0.7 & 101.80$_{\scriptstyle \pm 3.9}$ & 90.49$_{\scriptstyle \pm 4.84}$ & 64.78$_{\scriptstyle \pm 1.21}$ & \cellcolor{gray!15}\textbf{62.44$_{\scriptstyle \pm 0.27}$} & 50.93$_{\scriptstyle \pm 1.62}$ & 44.30$_{\scriptstyle \pm 2.26}$ & 46.45$_{\scriptstyle \pm 0.91}$ & \cellcolor{gray!15}\textbf{40.34$_{\scriptstyle \pm 0.38}$} \\
 & \multirow{2}{*}{8B} & 0.5 & 14.90$_{\scriptstyle \pm 0.48}$ & 13.20$_{\scriptstyle \pm 0.03}$ & 12.91$_{\scriptstyle \pm 0.06}$ & \cellcolor{gray!15}\textbf{11.90$_{\scriptstyle \pm 0.09}$} & 15.09$_{\scriptstyle \pm 0.07}$ & 12.61$_{\scriptstyle \pm 0.12}$ & 13.08$_{\scriptstyle \pm 0.09}$ & \cellcolor{gray!15}\textbf{12.10$_{\scriptstyle \pm 0.14}$} \\
 &  & 0.7 & 71.10$_{\scriptstyle \pm 1.86}$ & 65.46$_{\scriptstyle \pm 1.51}$ & 57.31$_{\scriptstyle \pm 1.28}$ & \cellcolor{gray!15}\textbf{52.56$_{\scriptstyle \pm 1.62}$} & 30.05$_{\scriptstyle \pm 0.66}$ & 25.13$_{\scriptstyle \pm 1.07}$ & 28.93$_{\scriptstyle \pm 0.43}$ & \cellcolor{gray!15}\textbf{20.10$_{\scriptstyle \pm 0.14}$} \\
 & \multirow{2}{*}{14B} & 0.5 & 11.43$_{\scriptstyle \pm 0.04}$ & 11.38$_{\scriptstyle \pm 0.04}$ & 10.79$_{\scriptstyle \pm 0.04}$ & \cellcolor{gray!15}\textbf{10.02$_{\scriptstyle \pm 0.03}$} & 11.51$_{\scriptstyle \pm 0.05}$ & 11.36$_{\scriptstyle \pm 0.02}$ & 10.92$_{\scriptstyle \pm 0.10}$ & \cellcolor{gray!15}\textbf{9.95$_{\scriptstyle \pm 0.13}$} \\
 &  & 0.7 & 37.04$_{\scriptstyle \pm 0.52}$ & 35.62$_{\scriptstyle \pm 0.27}$ & 34.29$_{\scriptstyle \pm 0.65}$ & \cellcolor{gray!15}\textbf{33.71$_{\scriptstyle \pm 0.45}$} & 20.82$_{\scriptstyle \pm 0.35}$ & 20.18$_{\scriptstyle \pm 0.16}$ & 20.37$_{\scriptstyle \pm 0.23}$ & \cellcolor{gray!15}\textbf{18.49$_{\scriptstyle \pm 0.28}$} \\
\bottomrule
\end{tabular}
\label{tab:ppl}
\vspace{-0.5cm}

\end{table*}

\begin{table}[t]
\centering
\small
\setlength{\tabcolsep}{6pt}
\caption{ \small Ablation study on hierarchical structural deletion. We report the absolute WikiText perplexity (PPL $\downarrow$)
of SWA and the increase in PPL for each ablated variant
relative to SWA on Qwen3 8B at sparsity levels of
0.5 and 0.7. SWA(Layer) uses only layer-level scores, and SWA(Head) uses only head-level scores.}
\label{tab:ablation_hier}
\resizebox{0.45\linewidth}{!}{

\begin{tabular}{lcccc}
\toprule
\multirow{2}{*}{\textbf{Variant}} 
& \multicolumn{2}{c}{\textbf{SparseGPT}} 
& \multicolumn{2}{c}{\textbf{Wanda}} \\
\cmidrule(lr){2-3} \cmidrule(lr){4-5}
& \textbf{0.5} & \textbf{0.7} 
& \textbf{0.5} & \textbf{0.7} \\
\midrule
\textbf{SWA} 
& \textbf{12.10} & \textbf{20.10} 
& \textbf{11.90} & \textbf{52.56} \\
SWA(Layer) 
& $+3.00$ & $+10.21$ & $+0.08$ & $+5.73$ \\
SWA(Head) 
& $+3.69$ & $+19.97$ & $+0.17$ & $+44.99$ \\
\bottomrule
\end{tabular}
}
\vspace{-0.5cm}
\end{table}


We conduct pruning experiments to validate our hypothesis
that attention heads with high core scores and low bridge
scores play an important role in supporting LLM reasoning
performance.
Specifically, we evaluate whether SWA better preserves both
small-world functional organization and model performance during pruning.
Our evaluation is guided by four research questions:
\textbf{RQ1.} Does SWA better preserve small-world functional
organization than existing pruning baselines?
\textbf{RQ2.} Does SWA better maintain model performance
than existing pruning baselines at the same sparsity level?
\textbf{RQ3.} How do layer-level and head-level sparsity
allocation contribute to SWA's effectiveness?
\textbf{RQ4.} How sensitive is SWA's pruning performance to the dataset used for functional graph construction?


\vspace{-0.2cm}
\subsection{Experimental Setup}
\label{sec:exp_setup}

First, we describe the setup used to collect activations, construct attention-head functional graphs, and compute SWI. Second, we describe the pruning setup used to evaluate whether SWA preserves performance and reasoning ability under sparsity. Details are provided in Appendix~\ref{app:expdetails}.

\paragraph{Setup for SWI analysis.}
We first evaluate the relationship between SWI and reasoning performance using six open-source LLMs: Qwen3-4B, Qwen3-8B, Qwen3-14B, Llama-3.2-1B, Llama-3.2-3B, and Meta-Llama-3-8B \citep{yang2025qwen3, grattafiori2024llama}. We collect attention-head activations on DRE-Bench \citep{yang2025truly}, using Level 1 and 2 questions as calibration inputs for constructing functional graphs. We exclude Levels 3 and 4 because current open-source models often fail on these harder tasks, making their answers unreliable for activation-based analysis. We measure DRE-Bench performance as the mean per-token negative log-likelihood of answer tokens (answer NLL/token; lower is better) and compare it with each model's SWI.

\paragraph{Setup for pruning evaluation.}
For pruning experiments, we use the same six LLMs and apply SWA to SparseGPT \citep{frantar2023sparsegpt} and Wanda \citep{sun2023simple}. We compare against uniform sparsity and layer-wise allocation baselines, including FARMS \citep{hu2025eigenspectrum} and ATP \citep{huang2025determining}. We evaluate pruned models using WikiText perplexity \citep{merity2016pointer}.

\subsection{Effectiveness of Preserving SWI}

To answer \textbf{RQ1}, we evaluate whether SWA preserves small-world functional organization during pruning. For each of the six models, we apply SparseGPT at a sparsity level of $0.5$ and compare it with variants using our score-guided allocation strategies. We then recompute the SWI of the resulting head--head functional graphs. \textbf{SWA (Head)} and \textbf{SWA (Layer)} denote head-only and layer-only allocation, respectively. As shown in Table~\ref{tab:swi_pruning_preservation}, SWA consistently achieves higher SWI values than the baseline pruning methods. Although the absolute SWI decreases substantially after pruning, SWA retains higher relative SWI than competing allocation strategies under the same sparsity level.

\subsection{Effectiveness of SWA}
To answer \textbf{RQ2}, we evaluate SWA on the WikiText validation set and report perplexity (PPL) in Table~\ref{tab:ppl}. SWA consistently outperforms the baselines across all models and sparsity levels, demonstrating that preserving small-world functional organization during pruning improves performance. Notably, the gap widens at higher sparsity, indicating that SWI-preserving pruning is particularly beneficial under aggressive pruning. Zero-shot results on downstream tasks are reported in Section~\ref{app:zero_shot_table} of the Appendix, and results on preserving reasoning ability on other benchmarks in Section~\ref{app:arauc}.

\subsection{Ablation of Hierarchical Sparsity Allocation}

To answer \textbf{RQ3}, we conduct an ablation study to examine the contribution of each level of SWA's hierarchical allocation. We compare our method with two ablated variants: \textbf{SWA(Layer)}, which uses only the layer-level scores, and \textbf{SWA(Head)}, which uses only the head-level scores. We evaluate all variants on Qwen3 8B with both SparseGPT and Wanda as the base pruning methods, using WikiText perplexity at sparsity levels of 0.5 and 0.7. As shown in Table~\ref{tab:ablation_hier}, removing either level increases perplexity in every setting, indicating that both head-level and layer-level structural signals contribute to preserving model quality after pruning. This degradation grows markedly from 0.5 to 0.7 sparsity for both base methods. Moreover, SWA(Head) degrades more than SWA(Layer) in every setting. 
These results suggest that layer-level scores provide the main allocation signal, while head-level scores refine pruning within layers.

\vspace{-0.2cm}
\subsection{Sensitivity to the Graph-Construction Dataset}

\begin{table}[t]
\centering
\caption{
Effect of the graph-construction benchmark on SWA.
We report WikiText perplexity ($\downarrow$) for
SparseGPT at 70\% sparsity, comparing the base method
with SWA using functional graphs built from GSM8K,
ARC-C, or MMLU.
}
\label{tab:graph_source_pruning}
\small
\setlength{\tabcolsep}{10pt}
\resizebox{0.7\linewidth}{!}{

\begin{tabular}{@{}lcccc@{}}
\toprule

\textbf{Model} & \textbf{Base} & \textbf{GSM8K} & \textbf{ARC-C} & \textbf{MMLU} \\
\midrule
Llama3 8B     &   55.94 $_{\scriptstyle \pm 3.23}$   &   42.60 $_{\scriptstyle \pm 1.22}$    & 41.10$_{\scriptstyle \pm 1.35}$      & 40.75 $_{\scriptstyle \pm 1.07}$      \\
Qwen3 4B      & 50.93 $_{\scriptstyle \pm 1.62}$ & 40.42 $_{\scriptstyle \pm 1.12}$ & 38.90 $_{\scriptstyle \pm 1.23}$ & 39.89$_{\scriptstyle \pm 2.35}$ \\

Llama3.2 3B   & 70.70 $_{\scriptstyle \pm 5.07}$ & 50.54$_{\scriptstyle \pm 1.30}$ & 49.69 $_{\scriptstyle \pm 2.69}$ & 48.95 $_{\scriptstyle \pm 1.20}$ \\
Llama3.2 1B   & 157.20$_{\scriptstyle \pm 3.20}$      & 109.2 $_{\scriptstyle \pm 6.20}$      & 110.9 $_{\scriptstyle \pm 2.60}$      & 105.28$_{\scriptstyle \pm 6.62}$      \\
\bottomrule
\end{tabular}
}
\end{table}

To answer \textbf{RQ4} we examine whether SWA remains effective when functional
graphs are constructed from different benchmarks.
Table~\ref{tab:graph_source_pruning} reports WikiText perplexity
for SparseGPT at 70\% sparsity, with SWA using graphs
constructed from GSM8K \citep{cobbe2021training}, ARC-C \citep{clark2018think}, or MMLU \citep{hendrycks2020measuring}. For each benchmark, we construct the functional graph using
400 training examples, or all available training examples
if fewer than 400 are available.
Across the four models with reported results, all three
graph sources yield lower perplexity than the base method.
For example, on Llama-3.2-3B, SWA reduces perplexity
from 70.70 to  48.95–50.54 across the three graph sources.
Although the best graph source varies across models,
these consistent improvements suggest that the benefits
of SWA are not limited to a single graph-construction
benchmark.



\vspace{-0.3cm}
\section{Related Works}
\vspace{-0.3cm}

\paragraph{Structural Analysis of Neural Networks.}
Prior work has studied neural-network structure through architecture graphs, weight spectra, and representation similarity. Architecture-level analyses relate graph properties such as clustering and path length to predictive performance, while weight-space methods based on random matrix theory connect spectral structure to training quality, generalization, and pruning sensitivity~\citep{you2020graph, martin2021implicit}. Representation-similarity methods such as SVCCA and CKA compare hidden states across layers and models, but do not explicitly model interactions among model components~\citep{raghu2017svcca, kornblith2019similarity, liu2025spectral}. For transformer-based models, prior studies show that attention heads are heterogeneous: some encode positional, syntactic, or induction-like behavior, while many are redundant or weakly used~\citep{clark2019does, voita2019analyzing, michel2019sixteen}. LLM pruning methods exploit this heterogeneity through weight- or layer-level importance estimates, including SparseGPT, Wanda, and ATP~\citep{frantar2023sparsegpt, sun2023simple, huang2025determining}. In contrast, our work constructs activation-defined functional graphs over attention heads, quantifies their global small-world organization, and derives local roles from this graph structure.

\paragraph{Functional Connectivity and Small-world Organization.}
Functional connectivity measures statistical dependence among distributed neural activities~\citep{friston1994functional}. Graph-theoretic work characterizes small-world networks by high clustering and short path lengths~\citep{watts1998collective}, or by high local and global efficiency~\citep{latora2001efficient}. In neuroscience, small-world organization is viewed as a balance between segregated specialization and integrated communication in brain networks~\citep{bassett2006small, bullmore2009complex}. Related machine-learning work uses activation-based graphs, activation correlations, and topological summaries to study neural-network organization and generalization~\citep{horta2021extracting, ballester2024predicting, ballester2023topological}, while transformer representation analyses reveal clustered structure across layers~\citep{phang2021fine}. Building on this perspective, we use SWI to test whether attention-head activations in LLMs form small-world functional organization associated with reasoning capability.

\section{Conclusion}

In this work, we investigated small-world functional organization in
attention-head networks and its relationship to LLM reasoning
performance. Across models and training checkpoints, higher
SWI is consistently associated with better fluid reasoning
performance. Our head-level analysis further reveals that
important heads tend to exhibit stronger within-community
connectivity and more concentrated connectivity across
communities, characterized by high core scores and low bridge
scores, respectively. Motivated by these findings, we proposed
Small-World Allocation (SWA) to examine whether preferentially
preserving heads with these properties helps maintain
performance under pruning. Experiments show that SWA better
preserves both small-world organization and model performance
than competing allocation strategies. Together, these findings
highlight the relevance of global functional organization and
local connectivity patterns to LLM reasoning, and demonstrate
their practical value for guiding model compression.

\subsection*{AI use statement}
We used generative AI tools to help with writing, editing, and proofreading this manuscript. The authors reviewed and checked all AI-assisted text, and take full responsibility for the content, methods, results, and conclusions of this paper.

\subsection*{Ethics statement}
This work uses publicly available models and datasets and
does not involve human-subject research or the collection
of personal or sensitive data. Our experiments focus on
analyzing attention-head functional organization and
evaluating model pruning. We do not identify additional
ethical concerns specific to the data or experimental
procedures used in this study.

\subsection*{Reproducibility statement}
We provide the methodological and experimental details
needed to reproduce our results. Sections~\ref{sec:prelim} and~\ref{key_findings} describe
the representation similarity measure, functional graph
construction, and small-world index. Section~\ref{sec:exp_setup} and
Appendix~\ref{app:expdetails} describe the models, datasets, evaluation
protocols, and pruning setup. Additional results are
provided in the appendix, and all models and datasets
used in our experiments are publicly available. We will release our codes to facilitate reproduction and further research.

\bibliography{iclr2027_conference}

@article{raven1962advanced,
  title={Advanced progressive matrices: Sets I and II},
  author={Raven, John C},
  journal={(No Title)},
  year={1962}
}

@article{watts1998collective,
  title={Collective dynamics of ‘small-world’networks},
  author={Watts, Duncan J and Strogatz, Steven H},
  journal={nature},
  volume={393},
  number={6684},
  pages={440--442},
  year={1998},
  publisher={Nature Publishing Group}
}

@article{wang2024mmlu,
  title={Mmlu-pro: A more robust and challenging multi-task language understanding benchmark},
  author={Wang, Yubo and Ma, Xueguang and Zhang, Ge and Ni, Yuansheng and Chandra, Abhranil and Guo, Shiguang and Ren, Weiming and Arulraj, Aaran and He, Xuan and Jiang, Ziyan and others},
  journal={Advances in Neural Information Processing Systems},
  volume={37},
  pages={95266--95290},
  year={2024}
}

@article{liu2025deepseek,
  title={Deepseek-v3. 2: Pushing the frontier of open large language models},
  author={Liu, Aixin and Mei, Aoxue and Lin, Bangcai and Xue, Bing and Wang, Bingxuan and Xu, Bingzheng and Wu, Bochao and Zhang, Bowei and Lin, Chaofan and Dong, Chen and others},
  journal={arXiv preprint arXiv:2512.02556},
  year={2025}
}

@article{leon2025gpt,
  title={GPT-5 and open-weight large language models: Advances in reasoning, transparency, and control},
  author={Leon, Maikel},
  journal={Information Systems},
  pages={102620},
  year={2025},
  publisher={Elsevier}
}

@article{bors2003effect,
  title={The effect of practice on Raven's Advanced Progressive Matrices},
  author={Bors, Douglas A and Vigneau, Fran{\c{c}}ois},
  journal={Learning and Individual Differences},
  volume={13},
  number={4},
  pages={291--312},
  year={2003},
  publisher={Elsevier}
}

@article{bassett2017small,
  title={Small-world brain networks revisited},
  author={Bassett, Danielle S and Bullmore, Edward T},
  journal={The Neuroscientist},
  volume={23},
  number={5},
  pages={499--516},
  year={2017},
  publisher={Sage Publications Sage CA: Los Angeles, CA}
}

@article{raghu2017svcca,
  title={Svcca: Singular vector canonical correlation analysis for deep learning dynamics and interpretability},
  author={Raghu, Maithra and Gilmer, Justin and Yosinski, Jason and Sohl-Dickstein, Jascha},
  journal={Advances in neural information processing systems},
  volume={30},
  year={2017}
}

@article{bullmore2009complex,
  title={Complex brain networks: graph theoretical analysis of structural and functional systems},
  author={Bullmore, Ed and Sporns, Olaf},
  journal={Nature reviews neuroscience},
  volume={10},
  number={3},
  pages={186--198},
  year={2009},
  publisher={Nature Publishing Group UK London}
}

@inproceedings{wu2025understanding,
  title={Understanding llms’ fluid intelligence deficiency: An analysis of the arc task},
  author={Wu, Junjie and Yu, Mo and Liu, Lemao and Yeung, Dit-Yan and Zhou, Jie},
  booktitle={Proceedings of the 2025 Conference of the Nations of the Americas Chapter of the Association for Computational Linguistics: Human Language Technologies (Volume 1: Long Papers)},
  pages={8339--8360},
  year={2025}
}

@inproceedings{mihaylov2018can,
  title={Can a suit of armor conduct electricity? a new dataset for open book question answering},
  author={Mihaylov, Todor and Clark, Peter and Khot, Tushar and Sabharwal, Ashish},
  booktitle={Proceedings of the 2018 conference on empirical methods in natural language processing},
  pages={2381--2391},
  year={2018}
}

@article{clark2018think,
  title={Think you have solved question answering? try arc, the ai2 reasoning challenge},
  author={Clark, Peter and Cowhey, Isaac and Etzioni, Oren and Khot, Tushar and Sabharwal, Ashish and Schoenick, Carissa and Tafjord, Oyvind},
  journal={arXiv preprint arXiv:1803.05457},
  year={2018}
}

@article{sakaguchi2021winogrande,
  title={Winogrande: An adversarial winograd schema challenge at scale},
  author={Sakaguchi, Keisuke and Bras, Ronan Le and Bhagavatula, Chandra and Choi, Yejin},
  journal={Communications of the ACM},
  volume={64},
  number={9},
  pages={99--106},
  year={2021},
  publisher={ACM New York, NY, USA}
}

@inproceedings{zellers2019hellaswag,
  title={Hellaswag: Can a machine really finish your sentence?},
  author={Zellers, Rowan and Holtzman, Ari and Bisk, Yonatan and Farhadi, Ali and Choi, Yejin},
  booktitle={Proceedings of the 57th annual meeting of the association for computational linguistics},
  pages={4791--4800},
  year={2019}
}

@inproceedings{wang2018glue,
  title={GLUE: A multi-task benchmark and analysis platform for natural language understanding},
  author={Wang, Alex and Singh, Amanpreet and Michael, Julian and Hill, Felix and Levy, Omer and Bowman, Samuel},
  booktitle={Proceedings of the 2018 EMNLP workshop BlackboxNLP: Analyzing and interpreting neural networks for NLP},
  pages={353--355},
  year={2018}
}

@inproceedings{clark2019boolq,
  title={Boolq: Exploring the surprising difficulty of natural yes/no questions},
  author={Clark, Christopher and Lee, Kenton and Chang, Ming-Wei and Kwiatkowski, Tom and Collins, Michael and Toutanova, Kristina},
  booktitle={Proceedings of the 2019 conference of the north American chapter of the association for computational linguistics: Human language technologies, volume 1 (long and short papers)},
  pages={2924--2936},
  year={2019}
}

@article{merity2016pointer,
  title={Pointer sentinel mixture models},
  author={Merity, Stephen and Xiong, Caiming and Bradbury, James and Socher, Richard},
  journal={arXiv preprint arXiv:1609.07843},
  year={2016}
}

@article{grattafiori2024llama,
  title={The llama 3 herd of models},
  author={Grattafiori, Aaron and Dubey, Abhimanyu and Jauhri, Abhinav and Pandey, Abhinav and Kadian, Abhishek and Al-Dahle, Ahmad and Letman, Aiesha and Mathur, Akhil and Schelten, Alan and Vaughan, Alex and others},
  journal={arXiv preprint arXiv:2407.21783},
  year={2024}
}

@article{yang2025qwen3,
  title={Qwen3 technical report},
  author={Yang, An and Li, Anfeng and Yang, Baosong and Zhang, Beichen and Hui, Binyuan and Zheng, Bo and Yu, Bowen and Gao, Chang and Huang, Chengen and Lv, Chenxu and others},
  journal={arXiv preprint arXiv:2505.09388},
  year={2025}
}

@article{martin2021predicting,
  title={Predicting trends in the quality of state-of-the-art neural networks without access to training or testing data},
  author={Martin, Charles H and Peng, Tongsu and Mahoney, Michael W},
  journal={Nature Communications},
  volume={12},
  number={1},
  pages={4122},
  year={2021},
  publisher={Nature Publishing Group UK London}
}

@article{thompson2000canonical,
  title={Canonical correlation analysis.},
  author={Thompson, Bruce},
  year={2000},
  publisher={American Psychological Association}
}

@article{blondel2008fast,
  title={Fast unfolding of communities in large networks},
  author={Blondel, Vincent D and Guillaume, Jean-Loup and Lambiotte, Renaud and Lefebvre, Etienne},
  journal={Journal of statistical mechanics: theory and experiment},
  volume={2008},
  number={10},
  pages={P10008},
  year={2008}
}

@article{yang2025truly,
  title={Truly assessing fluid intelligence of large language models through dynamic reasoning evaluation},
  author={Yang, Yue and Chen, MingKang and Liu, Qihua and Hu, Mengkang and Chen, Qiguang and Zhang, Gengrui and Hu, Shuyue and Zhai, Guangtao and Qiao, Yu and Wang, Yu and others},
  journal={arXiv preprint arXiv:2506.02648},
  year={2025}
}

@article{fox2005human,
  title={The human brain is intrinsically organized into dynamic, anticorrelated functional networks},
  author={Fox, Michael D and Snyder, Abraham Z and Vincent, Justin L and Corbetta, Maurizio and Van Essen, David C and Raichle, Marcus E},
  journal={Proceedings of the National Academy of Sciences},
  volume={102},
  number={27},
  pages={9673--9678},
  year={2005},
  publisher={National Academy of Sciences}
}

@article{biswal1995functional,
  title={Functional connectivity in the motor cortex of resting human brain using echo-planar MRI},
  author={Biswal, Bharat and Zerrin Yetkin, F and Haughton, Victor M and Hyde, James S},
  journal={Magnetic resonance in medicine},
  volume={34},
  number={4},
  pages={537--541},
  year={1995},
  publisher={Wiley Online Library}
}

@article{langer2012functional,
  title={Functional brain network efficiency predicts intelligence},
  author={Langer, Nicolas and Pedroni, Andreas and Gianotti, Lorena RR and H{\"a}nggi, J{\"u}rgen and Knoch, Daria and J{\"a}ncke, Lutz},
  journal={Human brain mapping},
  volume={33},
  number={6},
  pages={1393--1406},
  year={2012},
  publisher={Wiley Online Library}
}

@article{bassett2006small,
  title={Small-world brain networks},
  author={Bassett, Danielle Smith and Bullmore, ED},
  journal={The neuroscientist},
  volume={12},
  number={6},
  pages={512--523},
  year={2006},
  publisher={Sage Publications Sage CA: Thousand Oaks, CA}
}

@article{latora2001efficient,
  title={Efficient behavior of small-world networks},
  author={Latora, Vito and Marchiori, Massimo},
  journal={Physical review letters},
  volume={87},
  number={19},
  pages={198701},
  year={2001},
  publisher={APS}
}

@article{lu2024alphapruning,
  title={Alphapruning: Using heavy-tailed self regularization theory for improved layer-wise pruning of large language models},
  author={Lu, Haiquan and Zhou, Yefan and Liu, Shiwei and Wang, Zhangyang and Mahoney, Michael W and Yang, Yaoqing},
  journal={Advances in neural information processing systems},
  volume={37},
  pages={9117--9152},
  year={2024}
}

@article{barbey2018network,
  title={Network neuroscience theory of human intelligence},
  author={Barbey, Aron K},
  journal={Trends in cognitive sciences},
  volume={22},
  number={1},
  pages={8--20},
  year={2018},
  publisher={Elsevier}
}

@article{rosenberg2020functional,
  title={Functional segregation of human brain networks across the lifespan: an exploratory analysis of static and dynamic resting-state functional connectivity},
  author={Rosenberg, Benjamin M and Mennigen, Eva and Monti, Martin M and Kaiser, Roselinde H},
  journal={Frontiers in neuroscience},
  volume={14},
  pages={561594},
  year={2020},
  publisher={Frontiers Media SA}
}

@article{sun2023simple,
  title={A simple and effective pruning approach for large language models},
  author={Sun, Mingjie and Liu, Zhuang and Bair, Anna and Kolter, J Zico},
  journal={arXiv preprint arXiv:2306.11695},
  year={2023}
}

@inproceedings{frantar2023sparsegpt,
  title={Sparsegpt: Massive language models can be accurately pruned in one-shot},
  author={Frantar, Elias and Alistarh, Dan},
  booktitle={International conference on machine learning},
  pages={10323--10337},
  year={2023},
  organization={PMLR}
}

@inproceedings{kornblith2019similarity,
  title={Similarity of neural network representations revisited},
  author={Kornblith, Simon and Norouzi, Mohammad and Lee, Honglak and Hinton, Geoffrey},
  booktitle={International conference on machine learning},
  pages={3519--3529},
  year={2019},
  organization={PMlR}
}

@article{michel2019sixteen,
  title={Are sixteen heads really better than one?},
  author={Michel, Paul and Levy, Omer and Neubig, Graham},
  journal={Advances in neural information processing systems},
  volume={32},
  year={2019}
}

@inproceedings{voita2019analyzing,
  title={Analyzing multi-head self-attention: Specialized heads do the heavy lifting, the rest can be pruned},
  author={Voita, Elena and Talbot, David and Moiseev, Fedor and Sennrich, Rico and Titov, Ivan},
  booktitle={Proceedings of the 57th annual meeting of the association for computational linguistics},
  pages={5797--5808},
  year={2019}
}

@inproceedings{clark2019does,
  title={What does BERT look at? an analysis of BERT’s attention},
  author={Clark, Kevin and Khandelwal, Urvashi and Levy, Omer and Manning, Christopher D},
  booktitle={Proceedings of the 2019 ACL workshop BlackboxNLP: analyzing and interpreting neural networks for NLP},
  pages={276--286},
  year={2019}
}

@article{martin2021implicit,
  title={Implicit self-regularization in deep neural networks: Evidence from random matrix theory and implications for learning},
  author={Martin, Charles H and Mahoney, Michael W},
  journal={Journal of Machine Learning Research},
  volume={22},
  number={165},
  pages={1--73},
  year={2021}
}

@inproceedings{phang2021fine,
  title={Fine-tuned transformers show clusters of similar representations across layers},
  author={Phang, Jason and Liu, Haokun and Bowman, Samuel},
  booktitle={Proceedings of the Fourth BlackboxNLP Workshop on Analyzing and Interpreting Neural Networks for NLP},
  pages={529--538},
  year={2021}
}

@article{ballester2023topological,
  title={Topological data analysis for neural network analysis: A comprehensive survey},
  author={Ballester, Rub{\'e}n and Casacuberta, Carles and Escalera, Sergio},
  journal={arXiv preprint arXiv:2312.05840},
  year={2023}
}

@article{ballester2024predicting,
  title={Predicting the generalization gap in neural networks using topological data analysis},
  author={Ballester, Ruben and Clemente, Xavier Arnal and Casacuberta, Carles and Madadi, Meysam and Corneanu, Ciprian A and Escalera, Sergio},
  journal={Neurocomputing},
  volume={596},
  pages={127787},
  year={2024},
  publisher={Elsevier}
}

@article{horta2021extracting,
  title={Extracting knowledge from deep neural networks through graph analysis},
  author={Horta, Vitor AC and Tiddi, Ilaria and Little, Suzanne and Mileo, Alessandra},
  journal={Future Generation Computer Systems},
  volume={120},
  pages={109--118},
  year={2021},
  publisher={Elsevier}
}

@article{friston1994functional,
  title={Functional and effective connectivity in neuroimaging: a synthesis},
  author={Friston, Karl J},
  journal={Human brain mapping},
  volume={2},
  number={1-2},
  pages={56--78},
  year={1994},
  publisher={Wiley Online Library}
}

@article{huang2025determining,
  title={Determining layer-wise sparsity for large language models through a theoretical perspective},
  author={Huang, Weizhong and Zhang, Yuxin and Zheng, Xiawu and Chao, Fei and Ji, Rongrong},
  journal={arXiv preprint arXiv:2502.14770},
  year={2025}
}

@inproceedings{you2020graph,
  title={Graph structure of neural networks},
  author={You, Jiaxuan and Leskovec, Jure and He, Kaiming and Xie, Saining},
  booktitle={International conference on machine learning},
  pages={10881--10891},
  year={2020},
  organization={PMLR}
}

@article{hu2025eigenspectrum,
  title={Eigenspectrum analysis of neural networks without aspect ratio bias},
  author={Hu, Yuanzhe and Goel, Kinshuk and Killiakov, Vlad and Yang, Yaoqing},
  journal={arXiv preprint arXiv:2506.06280},
  year={2025}
}

@article{pedersen2017spontaneous,
  title={Spontaneous brain network activity: Analysis of its temporal complexity},
  author={Pedersen, Mangor and Omidvarnia, Amir and Walz, Jennifer M and Zalesky, Andrew and Jackson, Graeme D},
  journal={Network Neuroscience},
  volume={1},
  number={2},
  pages={100--115},
  year={2017},
  publisher={MIT Press One Rogers Street, Cambridge, MA 02142-1209, USA journals-info~…}
}

@article{hosseini2013influence,
  title={Influence of choice of null network on small-world parameters of structural correlation networks},
  author={Hosseini, SM Hadi and Kesler, Shelli R},
  journal={PLoS One},
  volume={8},
  number={6},
  pages={e67354},
  year={2013},
  publisher={Public Library of Science San Francisco, USA}
}

@inproceedings{biderman2023pythia,
  title={Pythia: A suite for analyzing large language models across training and scaling},
  author={Biderman, Stella and Schoelkopf, Hailey and Anthony, Quentin Gregory and Bradley, Herbie and O’Brien, Kyle and Hallahan, Eric and Khan, Mohammad Aflah and Purohit, Shivanshu and Prashanth, USVSN Sai and Raff, Edward and others},
  booktitle={International conference on machine learning},
  pages={2397--2430},
  year={2023},
  organization={PMLR}
}

@article{hendrycks2020measuring,
  title={Measuring massive multitask language understanding},
  author={Hendrycks, Dan and Burns, Collin and Basart, Steven and Zou, Andy and Mazeika, Mantas and Song, Dawn and Steinhardt, Jacob},
  journal={arXiv preprint arXiv:2009.03300},
  year={2020}
}

@article{holm1979simple,
  title={A simple sequentially rejective multiple test procedure},
  author={Holm, Sture},
  journal={Scandinavian journal of statistics},
  pages={65--70},
  year={1979},
  publisher={JSTOR}
}

@article{cobbe2021training,
  title={Training verifiers to solve math word problems},
  author={Cobbe, Karl and Kosaraju, Vineet and Bavarian, Mohammad and Chen, Mark and Jun, Heewoo and Kaiser, Lukasz and Plappert, Matthias and Tworek, Jerry and Hilton, Jacob and Nakano, Reiichiro and others},
  journal={arXiv preprint arXiv:2110.14168},
  year={2021}
}

@article{thompson2016stabilizing,
  title={On stabilizing the variance of dynamic functional brain connectivity time series},
  author={Thompson, William Hedley and Fransson, Peter},
  journal={Brain connectivity},
  volume={6},
  number={10},
  pages={735--746},
  year={2016},
  publisher={SAGE Publications Sage CA: Los Angeles, CA}
}

@article{whitfield2012conn,
  title={Conn: a functional connectivity toolbox for correlated and anticorrelated brain networks},
  author={Whitfield-Gabrieli, Susan and Nieto-Castanon, Alfonso},
  journal={Brain connectivity},
  volume={2},
  number={3},
  pages={125--141},
  year={2012},
  publisher={SAGE Publications Sage CA: Los Angeles, CA}
}

@inproceedings{liu2025spectral,
  title={Spectral insights into data-oblivious critical layers in large language models},
  author={Liu, Xuyuan and Hsiung, Lei and Yang, Yaoqing and Yan, Yujun},
  booktitle={Findings of the Association for Computational Linguistics: ACL 2025},
  pages={4860--4877},
  year={2025}
}
\bibliographystyle{iclr2027_conference}

\newpage
\appendix
\section{Appendix}

\subsection{Justification of the Head Similarity Measure}
\label{app:svcca_justification}

Our head similarity measure is a variant of SVCCA that
retains its SVD--CCA procedure and replaces mean
aggregation with the sum of Fisher $z$-transformed
canonical correlations. This formulation compares
vector-valued attention-head representations and
summarizes their shared variation across questions
as a scalar similarity score.

\paragraph{Canonical correlations for vector-valued activations.}
In functional brain networks, connectivity between
regions of interest (ROIs) is commonly estimated using
Pearson correlation between scalar ROI activity time
series~\citep{biswal1995functional,fox2005human}.
An attention head instead produces a multidimensional
activation vector, motivating a multivariate measure
of correlation. CCA identifies linear projections
of two representations that are maximally correlated,
as described in~\Cref{sec:svcca_prelim}.
For nonconstant one-dimensional representations,
the canonical correlation equals the absolute Pearson
correlation. CCA thus extends the correlation-based
perspective to vector-valued activations without
requiring coordinate-wise alignment.

\paragraph{SVD-based dimensionality reduction.}
Following the SVD--CCA procedure
of~\citet{raghu2017svcca}, we retain the leading
singular directions explaining at least 99\% of the
variance in each representation before applying CCA.
This step focuses the comparison on subspaces that
capture most of the observed variation and removes
low-variance directions that may reflect noise.
The resulting canonical correlations describe how
strongly the two heads' projected query activations
vary together across the sampled questions.

\paragraph{Fisher $z$-based aggregation.}
We apply the Fisher $z$-transform to each canonical
correlation and sum the transformed coefficients,
as defined in~\Cref{eq:fisher_similarity}.
The Fisher $z$-transform is widely used in functional
brain connectivity analysis to facilitate statistical
comparisons of correlation estimates
\citep{whitfield2012conn,thompson2016stabilizing}.
Here, we use its nonlinear scaling: for nonnegative
correlations, $\operatorname{atanh}(\rho)$ is increasing
and convex, placing greater emphasis on strongly
correlated canonical directions.
We clip the coefficients at $1-\epsilon_{\mathrm{clip}}$
to keep the score finite.
The sum measures aggregate alignment and depends on
both the correlation strengths and the number of
retained canonical directions.
We use this transformation as an aggregation rule,
without assuming that it stabilizes the sampling
variance of the estimated canonical correlations.

\paragraph{Comparison with CKA.}
Centered Kernel Alignment (CKA)~\citep{kornblith2019similarity}
compares centered similarity matrices over examples.
Our measure instead identifies correlated linear
projections and aggregates their correlation strengths.
We adopt this formulation to represent functional
connectivity between heads through multivariate
correlations, consistent with the correlation-based
perspective used in brain network analysis.
This choice reflects the intended interpretation of
our graph edges rather than a claim of general
superiority over CKA.

\subsection{Experimental Details}
\label{app:expdetails}

We first describe the datasets and settings used
for functional graph construction and SWI analysis.
We then present the pruning methods, allocation
baselines, and evaluation protocols.

\subsubsection{Graph Construction and SWI Analysis}
\label{sec:swi_analysis_details}

\paragraph{Graph-construction datasets.}
We use DRE-Bench~\citep{yang2025truly} for the main
SWI analysis and pruning experiments.
DRE-Bench evaluates fluid reasoning through abstract
tasks involving latent rules.
We sample 400 questions for graph construction:
200 from Level 1, which covers attribute-based
rules, and 200 from Level 2, which covers spatial
reasoning rules.

For the graph-construction dataset sensitivity
experiment, we additionally use
GSM8K~\citep{cobbe2021training},
ARC-C~\citep{clark2018think}, and
MMLU~\citep{hendrycks2020measuring}.
For each dataset, we use 400 training examples,
or all available training examples if fewer than
400 are available.
The examples used for graph construction are
excluded from downstream evaluations.

\paragraph{Activation collection and head similarity.}
We follow the activation collection procedure
in \Cref{sec:constructing-head}, recording query activations and
averaging them over the generated answer tokens
for each question.
We mean-center each head's activation matrix
and apply SVD, retaining the leading directions
that explain at least 99\% of its variance.
We then apply CCA to each pair of reduced
representations and aggregate the canonical
correlations using the Fisher $z$-based similarity
in~\Cref{eq:fisher_similarity}.
The variance threshold is shared across all models
and reported SWI analyses.

\paragraph{Graph construction and SWI computation.}
We construct graphs at a fixed edge density
$\delta=0.25$.
We rank all unordered pairs of distinct heads
by their Fisher $z$-aggregated similarity and
retain the top
$K_\delta=\lfloor\delta|\mathcal{P}|\rfloor$
pairs as edges, where $\mathcal{P}$ is the set
of eligible head pairs.
Self-connections are excluded.
We select $\delta=0.25$ based on a density sweep,
which identifies it as a consistent setting for
comparing SWI across model families.
We compute SWI following~\Cref{eq:swi}, using
degree-preserving randomized graphs as the reference.

\paragraph{Performance measurement.}
We measure DRE-Bench performance using the mean
per-token negative log-likelihood over the reference
answer tokens, where lower values indicate better
performance.
We compare this measure with SWI across the six
LLMs and across training checkpoints of Pythia-6.9B.

\subsubsection{Pruning Evaluation}
\label{sec:pruning_eval_details}

We integrate SWA with SparseGPT and Wanda and
compare it against the base methods with uniform
sparsity and variants using FARMS or ATP for
layer-wise sparsity allocation.

\paragraph{SparseGPT.}
SparseGPT~\citep{frantar2023sparsegpt} is a one-shot
post-training pruning method that approximates
each layer's original outputs on calibration data.
It uses second-order information to select pruning
masks and update the remaining weights without
fine-tuning.
We replace its uniform sparsity setting with
the allocations produced by SWA or the competing
allocation methods.

\paragraph{Wanda.}
Wanda~\citep{sun2023simple} estimates weight importance
using weight magnitude and input activation statistics.
For a weight matrix $W$, it scores each weight by
$|W_{ij}|\cdot\|X_j\|_2$, where $X_j$ contains
the observed activations of input feature $j$.
Weights with lower scores are pruned within each
output channel.
Wanda does not update the remaining weights after
pruning. We use it as a second base method to
evaluate SWA beyond reconstruction-based pruning.

\paragraph{FARMS.}
FARMS~\citep{hu2025eigenspectrum} is an eigenspectrum-based
layer-wise allocation method that addresses
aspect-ratio bias in heavy-tailed self-regularization
analysis.
It samples submatrices with a fixed aspect ratio
and aggregates their empirical spectral densities
to obtain comparable estimates across heterogeneous
weight matrices.
These estimates guide layer-wise sparsity allocation.

\paragraph{ATP.}
ATP~\citep{huang2025determining} allocates layer-wise
sparsity to mitigate the propagation of reconstruction
errors during pruning.
It assigns lower sparsity to earlier layers and
higher sparsity to later layers using an increasing
arithmetic progression, with the common difference
selected through search.

\paragraph{Language modeling and zero-shot evaluation.}
We evaluate language modeling using WikiText
perplexity~\citep{merity2016pointer}, where lower
values indicate better performance.
We also report average zero-shot accuracy across
BoolQ~\citep{clark2019boolq},
RTE~\citep{wang2018glue},
HellaSwag~\citep{zellers2019hellaswag},
WinoGrande~\citep{sakaguchi2021winogrande},
ARC-Easy and ARC-Challenge~\citep{clark2018think},
and OpenBookQA~\citep{mihaylov2018can}.
We evaluate both base pruning methods at sparsity
levels of 0.5 and 0.7.
Results are averaged over three seeds and reported
as mean$_{\pm\mathrm{std}}$.

\paragraph{ARAOC evaluation.}
We additionally evaluate pruned models on the
Abstraction and Reasoning on Atom Operation Corpus
(ARAOC)~\citep{wu2025understanding}.
ARAOC contains six types of atomic grid transformations:
\textit{Move}, \textit{Change Color}, \textit{Copy},
\textit{Mirror}, \textit{Fill Internal}, and
\textit{Scale}.

We report \textbf{Not M}, the percentage of predictions
whose output-grid shape differs from the ground-truth
shape. Lower values indicate better preservation
of output-shape correctness.
ARAOC is used exclusively for evaluation and does
not contribute to functional graph construction.
We evaluate Qwen3-8B pruned with SparseGPT at 50\%
sparsity and report the mean and standard deviation
over three runs.

\subsubsection{Computing Infrastructure}

All experiments are conducted on a cluster equipped
with NVIDIA RTX PRO 6000 GPUs and AMD EPYC 9655
processors.

\subsection{Numerical Results for the SWI--Performance Association}
\label{app:add_swi_fluid}

\begin{table}[t]
\centering
\small
\setlength{\tabcolsep}{4pt}

\caption{\small
Relationship between DRE-Bench performance and small-world
organization across Pythia-6.9B training checkpoints.
We report the mean per-token negative log-likelihood over
the answer tokens (NLL/token; lower is better) and SWI of
the head-head functional graph on DRE-Bench Level 1 and
Level 2 tasks.
Results are averaged over three seeds and reported as
mean$_{\pm \text{std}}$.
Across training checkpoints, lower NLL is associated with
higher SWI on both levels.
}
\label{tab:pythia_checkpoints}

\begin{tabular}{rrcccc}
\toprule
\textbf{Training step} & \textbf{Tokens seen}
& \multicolumn{2}{c}{\textbf{Level 1}}
& \multicolumn{2}{c}{\textbf{Level 2}} \\
\cmidrule(lr){3-4} \cmidrule(lr){5-6}
& &
\textbf{NLL/token} $\downarrow$
& \textbf{SWI} $\uparrow$
& \textbf{NLL/token} $\downarrow$
& \textbf{SWI} $\uparrow$ \\
\midrule
4,000 & 8B
& 0.5089$_{\scriptstyle \pm 0.16}$
& 0.6760$_{\scriptstyle \pm 0.85}$
& 0.8255$_{\scriptstyle \pm 0.25}$
& 0.4587$_{\scriptstyle \pm 0.67}$ \\

11,000 & 23B
& 0.4677$_{\scriptstyle \pm 0.20}$
& 0.8008$_{\scriptstyle \pm 0.13}$
& 0.7020$_{\scriptstyle \pm 0.68}$
& 0.5574$_{\scriptstyle \pm 0.80}$ \\

30,000 & 63B
& 0.4344 $_{\scriptstyle \pm 0.28}$
& 0.8035 $_{\scriptstyle \pm 0.12}$
& 0.7007$_{\scriptstyle \pm 0.52}$
& 0.5599$_{\scriptstyle \pm 0.54}$
\\
65,000 & 136B
& 0.2250 $_{\scriptstyle \pm 0.30}$
& 0.8135 $_{\scriptstyle \pm 0.19}$
& 0.6411$_{\scriptstyle \pm 0.34}$
& 0.8100$_{\scriptstyle \pm 0.32}$
\\

71,000 & 149B
& 0.1717$_{\scriptstyle \pm 0.07}$
& 0.8320$_{\scriptstyle \pm 0.47}$
& 0.4026$_{\scriptstyle \pm 0.18}$
& 0.8469$_{\scriptstyle \pm 0.33}$ \\
113,000 & 237B
& 0.1623$_{\scriptstyle \pm 0.09}$
& 0.9768$_{\scriptstyle \pm 0.45}$
& 0.3765$_{\scriptstyle \pm 0.17}$
& 0.9996$_{\scriptstyle \pm 0.88}$ \\
143,000 & 300B
& 0.1374$_{\scriptstyle \pm 0.03}$
& 1.0075$_{\scriptstyle \pm 0.22}$
& 0.3749$_{\scriptstyle \pm 0.03}$
& 1.1864$_{\scriptstyle \pm 0.21}$ \\
\bottomrule
\end{tabular}
\vspace{-0.5cm}
\end{table}
\begin{table}[t]
\centering
\small
\setlength{\tabcolsep}{4pt}

\caption{\small Relationship between DRE-Bench performance and small-world organization.
We report the mean per-token negative log-likelihood over the answer tokens (NLL/token; lower is better) and SWI of the head-head functional graph on DRE-Bench Level 1 and Level 2 tasks. Results are averaged over three seeds and reported as mean$_{\pm \text{std}}$. Across models, lower NLL is associated with higher SWI.}

\begin{tabular}{lcccc}
\toprule
\textbf{Model} 
& \multicolumn{2}{c}{\textbf{Level 1}} 
& \multicolumn{2}{c}{\textbf{Level 2}} \\
\cmidrule(lr){2-3} \cmidrule(lr){4-5}
& \textbf{NLL/token} $\downarrow$
& \textbf{SWI} $\uparrow$
& \textbf{NLL/token} $\downarrow$
& \textbf{SWI} $\uparrow$ \\
\midrule
Qwen3 14B 
& 0.0822$_{\scriptstyle \pm 0.01}$ & 1.8160$_{\scriptstyle \pm 0.27}$
& 0.2410$_{\scriptstyle \pm 0.03}$ & 1.7020$_{\scriptstyle \pm 0.12}$ \\
Qwen3 8B 
& 0.0949$_{\scriptstyle \pm 0.01}$ & 1.5850$_{\scriptstyle \pm 0.28}$
& 0.2814$_{\scriptstyle \pm 0.02}$ & 1.6000$_{\scriptstyle \pm 0.07}$\\
Llama3 8B 
& 0.0980$_{\scriptstyle \pm 0.01}$ & 1.3750$_{\scriptstyle \pm 0.19}$
& 0.2993$_{\scriptstyle \pm 0.02}$ & 1.5935$_{\scriptstyle \pm 0.11}$ \\
Qwen3 4B 
& 0.1142$_{\scriptstyle \pm 0.02}$ & 1.2180$_{\scriptstyle \pm 0.26}$
& 0.3397$_{\scriptstyle \pm 0.05}$ & 1.2722$_{\scriptstyle \pm 0.18}$ \\
Llama3.2 3B 
& 0.1377$_{\scriptstyle \pm 0.07}$ & 1.1933$_{\scriptstyle \pm 0.33}$
& 0.3412$_{\scriptstyle \pm 0.10}$ & 1.2100$_{\scriptstyle \pm 0.17}$ \\
Llama3.2 1B 
& 0.1689$_{\scriptstyle \pm 0.20}$ & 0.9188$_{\scriptstyle \pm 0.49}$
& 0.5800$_{\scriptstyle \pm 0.26}$ & 0.9344$_{\scriptstyle \pm 0.18}$ \\
\bottomrule
\end{tabular}
\vspace{-0.5cm}

\label{tab:dre_perf_swi_levels}
\end{table}

Tables~\ref{tab:pythia_checkpoints}
and~\ref{tab:dre_perf_swi_levels} provide detailed
numerical results for the checkpoint and cross-model
analyses presented in
Figures~\ref{fig:swi-perf}
and~\ref{fig:swi-perf-cross_model}, respectively.
We report mean answer-token negative log-likelihood
(NLL/token) and SWI separately for DRE-Bench Level 1
and Level 2. Results are averaged over three seeds
and reported as mean$_{\pm\mathrm{std}}$.

\paragraph{Across training checkpoints.}
Table~\ref{tab:pythia_checkpoints} reports results
for seven training checkpoints of
Pythia-6.9B~\citep{biderman2023pythia}.
On both levels, mean answer-token NLL decreases
as training progresses, while mean SWI increases.
For example, on Level 2, NLL decreases from 0.8255
at step 4,000 to 0.3749 at step 143,000, while SWI
increases from 0.4587 to 1.1864.
This comparison examines the association across
checkpoints with the same architecture and parameter
count.

\paragraph{Across models.}
Table~\ref{tab:dre_perf_swi_levels} reports results
for six LLMs from the Qwen and Llama families.
On both levels, models with lower mean answer-token
NLL consistently exhibit higher mean SWI.
Qwen3-14B achieves the lowest NLL and highest SWI,
whereas Llama-3.2-1B exhibits the highest NLL and
lowest SWI. These numerical results document the
cross-model trend discussed
in~\Cref{sec:quantify_swi}.

\subsection{Zero-Shot Performance of SWA}
\label{app:zero_shot_table}

\begin{table*}[t]
\centering
\scriptsize
\setlength{\tabcolsep}{2.0pt}
\caption{\small Comparison of Zero-shot Accuracy  ($\uparrow$). We evaluate Base pruning methods (Wanda and SparseGPT) and their variants augmented with FARMS, ATP, and SWA at sparsity 0.5/0.7. Results are averaged over three seeds and reported as mean$_{\pm \text{std}}$. The best result in each setting is \colorbox{gray!15}{\textbf{bolded}}. SWA improves zero-shot accuracy and achieves the strongest performance over all the models and sparsity levels.}
\begin{tabular}{ccc|llll|llll}
\toprule
\multirow{2}{*}{\textbf{Model}} & \multirow{2}{*}{\textbf{Size}} & \multirow{2}{*}{\textbf{Sparsity}} & \multicolumn{4}{c|}{\textbf{Wanda}} & \multicolumn{4}{c}{\textbf{SparseGPT}} \\
 &  &  & Base & + FARMS & + ATP & \textbf{+ SWA} & Base & + FARMS & + ATP & \textbf{+ SWA} \\
\midrule
\multirow{4}{*}{\shortstack{Llama 3.2}} & \multirow{2}{*}{1B} & 0.5 & 34.91$_{\scriptstyle \pm 3.67}$ & 35.21$_{\scriptstyle \pm 2.28}$ & 37.16$_{\scriptstyle \pm 0.82}$ & \cellcolor{gray!15}\textbf{38.89$_{\scriptstyle \pm 1.03}$} & 39.57$_{\scriptstyle \pm 2.67}$ & 40.33$_{\scriptstyle \pm 2.41}$ & 41.26$_{\scriptstyle \pm 0.23}$ & \cellcolor{gray!15}\textbf{44.67$_{\scriptstyle \pm 1.75}$}\\
 &  & 0.7 & 25.86$_{\scriptstyle \pm 3.62}$ & 27.29$_{\scriptstyle \pm 4.75}$ & 29.86$_{\scriptstyle \pm 4.65}$ & \cellcolor{gray!15}\textbf{32.41$_{\scriptstyle \pm 4.81}$}& 30.85$_{\scriptstyle \pm 4.67}$ & 29.38$_{\scriptstyle \pm 1.33}$ & 35.29$_{\scriptstyle \pm 0.22}$ & \cellcolor{gray!15}\textbf{36.46$_{\scriptstyle \pm 0.34}$}\\
 & \multirow{2}{*}{3B} & 0.5 & 40.53$_{\scriptstyle \pm 1.41}$ & 42.71$_{\scriptstyle \pm 1.22}$ & 44.57$_{\scriptstyle \pm 1.39}$ & \cellcolor{gray!15}\textbf{45.80$_{\scriptstyle \pm 1.87}$}& 45.50$_{\scriptstyle \pm 0.36}$ & 46.67$_{\scriptstyle \pm 0.50}$ & 46.05$_{\scriptstyle \pm 1.41}$ & \cellcolor{gray!15}\textbf{49.91$_{\scriptstyle \pm 0.31}$}\\
 &  & 0.7 & 40.16$_{\scriptstyle \pm 2.41}$ & 41.50$_{\scriptstyle \pm 1.63}$ & 40.91$_{\scriptstyle \pm 4.74}$ & \cellcolor{gray!15}\textbf{43.04$_{\scriptstyle \pm 0.87}$} & 40.42$_{\scriptstyle \pm 1.16}$ & 42.25$_{\scriptstyle \pm 0.85}$ & 41.69$_{\scriptstyle \pm 4.46}$ & \cellcolor{gray!15}\textbf{43.57$_{\scriptstyle \pm 1.11}$} \\
\multirow{2}{*}{\shortstack{Llama 3}} & \multirow{2}{*}{8B} & 0.5 & 52.12$_{\scriptstyle \pm 1.25}$ & 53.49$_{\scriptstyle \pm 2.17}$ & 52.43$_{\scriptstyle \pm 1.32}$ & \cellcolor{gray!15}\textbf{55.13$_{\scriptstyle \pm 1.52}$}& 48.53$_{\scriptstyle \pm 2.24}$ & 50.28$_{\scriptstyle \pm 1.30}$ & 53.48$_{\scriptstyle \pm 0.61}$ & \cellcolor{gray!15}\textbf{55.61$_{\scriptstyle \pm 0.85}$}\\
 &  & 0.7 & 27.57$_{\scriptstyle \pm 0.07}$ & 27.14$_{\scriptstyle \pm 1.84}$ & 35.90$_{\scriptstyle \pm 0.74}$ & \cellcolor{gray!15}\textbf{37.89$_{\scriptstyle \pm 0.90}$}& 37.26$_{\scriptstyle \pm 5.39}$ & 38.38$_{\scriptstyle \pm 0.90}$ & 42.33$_{\scriptstyle \pm 2.86}$ & \cellcolor{gray!15}\textbf{43.32$_{\scriptstyle \pm 1.56}$} \\
\midrule
\multirow{6}{*}{\shortstack{Qwen 3}} & \multirow{2}{*}{4B} & 0.5 & 45.69$_{\scriptstyle \pm 3.35}$ & 46.33$_{\scriptstyle \pm 2.49}$ & 47.45$_{\scriptstyle \pm 0.39}$ & \cellcolor{gray!15}\textbf{48.07$_{\scriptstyle \pm 0.42}$}& 43.75$_{\scriptstyle \pm 1.25}$ & 47.28$_{\scriptstyle \pm 1.04}$ & 46.95$_{\scriptstyle \pm 4.51}$ & \cellcolor{gray!15}\textbf{48.00$_{\scriptstyle \pm 1.23}$} \\
 &  & 0.7 & 35.86$_{\scriptstyle \pm 0.06}$ & 36.60$_{\scriptstyle \pm 1.64}$ & 36.98$_{\scriptstyle \pm 0.48}$ & \cellcolor{gray!15}\textbf{38.30$_{\scriptstyle \pm 0.68}$}& 41.21$_{\scriptstyle \pm 4.51}$ & 42.50$_{\scriptstyle \pm 0.89}$ & 42.00$_{\scriptstyle \pm 4.02}$ & \cellcolor{gray!15}\textbf{43.02$_{\scriptstyle \pm 1.23}$} \\
 & \multirow{2}{*}{8B} & 0.5 & 52.50$_{\scriptstyle \pm 0.89}$ & 53.28$_{\scriptstyle \pm 0.77}$ & 55.38$_{\scriptstyle \pm 0.19}$ & \cellcolor{gray!15}\textbf{57.84$_{\scriptstyle \pm 0.24}$}& 53.04$_{\scriptstyle \pm 3.22}$ & 54.11$_{\scriptstyle \pm 2.70}$ & 55.64$_{\scriptstyle \pm 0.86}$ & \cellcolor{gray!15}\textbf{56.59$_{\scriptstyle \pm 0.68}$}\\
 &  & 0.7 & 38.95$_{\scriptstyle \pm 2.04}$ & 39.57$_{\scriptstyle \pm 0.53}$ & 39.64$_{\scriptstyle \pm 0.53}$ & \cellcolor{gray!15}\textbf{41.66$_{\scriptstyle \pm 0.88}$} & 42.24$_{\scriptstyle \pm 0.52}$ & 42.91$_{\scriptstyle \pm 1.72}$ & 43.48$_{\scriptstyle \pm 0.74}$ & \cellcolor{gray!15}\textbf{46.11$_{\scriptstyle \pm 0.72}$}\\
 & \multirow{2}{*}{14B} & 0.5 & 57.05$_{\scriptstyle \pm 0.44}$ & 57.31$_{\scriptstyle \pm 0.22}$ & 57.86$_{\scriptstyle \pm 1.15}$ & \cellcolor{gray!15}\textbf{59.30$_{\scriptstyle \pm 0.45}$}& 57.12$_{\scriptstyle \pm 0.39}$ & 57.62$_{\scriptstyle \pm 4.73}$ & 57.38$_{\scriptstyle \pm 0.15}$ & \cellcolor{gray!15}\textbf{58.10$_{\scriptstyle \pm 0.34}$} \\
 &  & 0.7 & 43.91$_{\scriptstyle \pm 3.11}$ & 44.69$_{\scriptstyle \pm 2.42}$ & 45.41$_{\scriptstyle \pm 0.93}$ & \cellcolor{gray!15}\textbf{46.23$_{\scriptstyle \pm 2.71}$}& 48.43$_{\scriptstyle \pm 2.22}$ & 49.11$_{\scriptstyle \pm 1.80}$ & 48.97$_{\scriptstyle \pm 2.01}$ & \cellcolor{gray!15}\textbf{49.93$_{\scriptstyle \pm 2.22}$} \\
\bottomrule
\end{tabular}
\label{tab:zero}
\end{table*}

Table~\ref{tab:zero} reports average zero-shot accuracy
on the downstream tasks described
in~\Cref{sec:pruning_eval_details}.
SWA achieves the highest mean accuracy across
all six models at both sparsity levels when
combined with either Wanda or SparseGPT.
At 70\% sparsity, SWA improves the average accuracy
of Wanda-pruned Llama-3-8B from 27.57\% to 37.89\%,
outperforming the strongest competing allocation
method, ATP, by 1.99 percentage points.
These results show that SWA's benefits extend
beyond language modeling perplexity to downstream
task performance.

\subsection{Performance on ARAOC Reasoning Tasks}
\label{app:arauc}

\begin{table*}[t]
\centering
\scriptsize
\setlength{\tabcolsep}{6pt}

\caption{\small ARAOC task mismatch rate at sparsity 0.5. We report mismatch rate (Not M $\downarrow$) across six tasks. Results are averaged over three runs and reported as mean$_{\pm \text{std}}$. The best result in each setting is \colorbox{gray!15}{\textbf{bolded}}.}

\begin{tabular}{l|cccccc}
\toprule
\textbf{Method} 
& \textbf{Move} 
& \textbf{Change Color} 
& \textbf{Copy} 
& \textbf{Mirror} 
& \textbf{Fill Internal} 
& \textbf{Scale} \\
\midrule

Qwen3 8B (Non-Pruned) 
& 36.00$_{\scriptstyle \pm 4.11}$
& 23.14$_{\scriptstyle \pm 4.61}$
& 12.18$_{\scriptstyle \pm 4.12}$
& 84.66$_{\scriptstyle \pm 3.20}$
& 6.07$_{\scriptstyle \pm 1.99}$
& 83.33$_{\scriptstyle \pm 5.87}$ \\
\midrule

SparseGPT 
& 44.21$_{\scriptstyle \pm 9.20}$
& 48.87$_{\scriptstyle \pm 10.40}$
& 60.99$_{\scriptstyle \pm 10.67}$
& 94.76$_{\scriptstyle \pm 8.21}$
& 34.25$_{\scriptstyle \pm 4.67}$
& 88.91$_{\scriptstyle \pm 8.03}$ \\

SparseGPT + FARMS 
& 46.22$_{\scriptstyle \pm 8.21}$
& 50.77$_{\scriptstyle \pm 15.85}$
& 56.60$_{\scriptstyle \pm 12.80}$
& 96.66$_{\scriptstyle \pm 5.48}$
& 33.11$_{\scriptstyle \pm 5.11}$
& 88.06$_{\scriptstyle \pm 8.95}$ \\

SparseGPT + ATP 
& 43.09$_{\scriptstyle \pm 9.01}$
& 47.73$_{\scriptstyle \pm 8.66}$
& 56.79$_{\scriptstyle \pm 12.48}$
& 93.56$_{\scriptstyle \pm 9.63}$
& 32.66$_{\scriptstyle \pm 5.81}$
& 89.49$_{\scriptstyle \pm 7.47}$ \\

SparseGPT + \textbf{SWA} 
& \cellcolor{gray!15}\textbf{40.29$_{\scriptstyle \pm 9.25}$}
& \cellcolor{gray!15}\textbf{46.28$_{\scriptstyle \pm 9.62}$}
& \cellcolor{gray!15}\textbf{56.11$_{\scriptstyle \pm 11.62}$}
& \cellcolor{gray!15}\textbf{90.21$_{\scriptstyle \pm 8.89}$}
& \cellcolor{gray!15}\textbf{28.69$_{\scriptstyle \pm 4.28}$}
& \cellcolor{gray!15}\textbf{85.10$_{\scriptstyle \pm 7.28}$} \\

\bottomrule
\end{tabular}
\vspace{-0.5cm}

\label{tab:araoc_split}
\end{table*}

Table~\ref{tab:araoc_split} reports ARAOC output-grid
shape mismatch rates for Qwen3-8B at 50\% sparsity.
The benchmark and evaluation metric are described
in~\Cref{sec:pruning_eval_details}.
SWA achieves lower mismatch rates than the pruning
baselines across all six task types.
For example, on \textit{Fill Internal}, SWA reduces
the mismatch rate from 34.25\% with SparseGPT to
28.69\%, compared with 32.66\% for ATP.
Although mismatch rates remain higher than those
of the unpruned model, SWA better preserves
output-shape correctness on these abstract
reasoning tasks.

\end{document}